\documentclass[journal]{IEEEtran}
\usepackage{amsmath,amsfonts}
\usepackage{algorithmicx}
\usepackage{algorithm}
\usepackage{array}
\usepackage[caption=false,font=normalsize,labelfont=sf,textfont=sf]{subfig}
\usepackage{textcomp}
\usepackage{stfloats}
\usepackage{url}
\usepackage{verbatim}
\usepackage{graphicx}
\usepackage{cite}
\usepackage{enumitem} 
\usepackage{xcolor} 
\usepackage{booktabs} 
\usepackage{tabularx} 
\usepackage{array} 
\usepackage{multirow} 
\usepackage{siunitx} 
\usepackage{threeparttable} 
\usepackage{ulem} 
\usepackage{hyperref} 
\usepackage{algpseudocode}
\begin{document}

\title{A Genetic Algorithm-Based Approach for Automated Optimization of Kolmogorov-Arnold Networks in Classification Tasks}

\author{
Quan~Long,~\IEEEmembership{Student Member,~IEEE,}
Bin~Wang,~\IEEEmembership{Member,~IEEE,}
Bing~Xue,~\IEEEmembership{Fellow,~IEEE,}
Mengjie~Zhang,~\IEEEmembership{Fellow,~IEEE}
\thanks{Quan Long is with Hebei University, Baoding, China. (e-mail: longquan@stumail.hbu.edu.cn).}
}



\maketitle

\begin{abstract}
To address the issue of interpretability in multilayer perceptrons (MLPs), Kolmogorov-Arnold Networks (KANs) are introduced in 2024. However, optimizing KAN structures is labor-intensive, typically requiring manual intervention and parameter tuning. 
This paper proposes GA-KAN, a genetic algorithm-based approach that automates the optimization of KANs, requiring no human intervention in the design process.
To the best of our knowledge, this is the first time that evolutionary computation is explored to optimize KANs automatically. 
Furthermore, inspired by the use of sparse connectivity in MLPs in effectively reducing the number of parameters, GA-KAN further explores sparse connectivity to tackle the challenge of extensive parameter spaces in KANs.
GA-KAN is validated on two toy datasets, achieving optimal results without the manual tuning required by the original KAN. 
Additionally, GA-KAN demonstrates superior performance across five classification datasets, outperforming traditional methods on all datasets and providing interpretable symbolic formulae for the Wine and Iris datasets, thereby enhancing model transparency.
Furthermore, GA-KAN significantly reduces the number of parameters over the standard KAN across all the five datasets. The core contributions of GA-KAN include automated optimization, a new encoding strategy, and a new decoding process, which together improve the accuracy and interpretability, and reduce the number of parameters. 
\end{abstract}


\section{Introduction}
\IEEEPARstart{A}{rtificial} neural networks (ANNs) have garnered significant attention for decades. ANNs are computational algorithms used to model data in scenarios where it is challenging to extract trends or detect patterns\cite{abdolrasol2021artificial}. They have made substantial advancements in various fields, including natural language processing\cite{khurana2023natural} and computer vision\cite{bhatt2021cnn}. Common ANN architectures \cite{yang2020artificial} include multilayer perceptrons (MLPs), recurrent neural networks (RNNs), and convolutional neural networks (CNNs). Among these, the MLP models are widely regarded as a powerful approach for modeling non-linear functions in machine learning. 
However, MLPs also have a couple of major limitations. First, as the number of layers or trainable parameters increases, the computational cost becomes significantly higher \cite{al2020autoencoder}. Second, MLPs lack interpretability, which can hinder their practical application in some contexts \cite{hakkoum2021assessing, zhang2021survey}.

To tackle the challenge of extensive parameter spaces in MLPs, Mocanu \textit{et al.} \cite{mocanu2018scalable} proposed a method that uses sparse layers to replace fully connected layers, demonstrating that it is feasible to significantly reduce the number of parameters while maintaining MLP's performance.
Consequently, exploring connectivity patterns within neural networks has become a promising direction for model optimization, crucial for enhancing efficiency, performance, and adaptability across various applications.

In 2024, Liu \textit{et al.} \cite{KAN} introduced a novel neural network architecture known as Kolmogorov-Arnold Networks (KANs) to improve interpretability as a promising alternative to MLPs. Unlike MLPs, which can fit the input-output relationships in data but typically produce implicit rather than explicit symbolic representations, KANs can capture symbolic expressions of input-output relationships. 
These capabilities provide KANs with potential interpretability that surpasses that of traditional MLPs. A key distinction between KANs and MLPs lies in their architecture: KANs employ learnable activation functions on the edges, while nodes are limited to simple summation operations. \cite{KAN} pointed out that the parameters of manually constructed KANs are often redundant, offering significant room for optimization. 

Sparse connectivity plays a crucial role in neural networks by reducing the number of parameters and enhancing efficiency. KANs have adopted pruning methods to reduce the network size \cite{KAN}. However, the pruning process requires manual tuning of parameters, which may vary across datasets and require expert knowledge. Additionally, the pruning method restricts the number of layers in KANs, limiting their flexibility in adapting to different datasets and preventing the network from dynamically adjusting its depth to optimize performance. Inspired by the use of sparse connectivity in MLPs, we explore sparse connections in KANs. 

In our approach, the connections between neurons and the network depth, along with the grid value, are encoded into vectors. Then, genetic algorithms (GAs) are applied to automate the optimization process for the following two major reasons. First of all, GA, known for its low computational cost and effective optimization strategies, is widely used in optimization and design \cite{albadr2019spoken}. In addition, GA-based Neural Architecture Search (NAS) has shown promising performance in recent studies \cite{feng2024enhancing}, making it the preferred choice for this task.
This approach aims to identify the optimal network architecture in KANs for given tasks and to address the limitations of manual parameter tuning in traditional pruning methods.

To summarize, the overall goal of this paper is to propose GA-KAN, a GA-based approach that explores sparse connections of KANs and automatically optimizes KAN architectures. To the best of our knowledge, this is the first time that evolutionary computation methods have been explored to automatically optimize KANs.
The overall goal is achieved through the following objectives:

\begin{enumerate}
    \item \textbf{Automated Optimization}: GA-KAN utilizes a GA to optimize both the KAN architecture and grid values through a unique encoding and decoding strategy. This enables GA-KAN to automatically identify the optimal KAN for superior performance in classification tasks, with no human intervention in the design process and minimal manual adjustment in the formula extraction process.
    \item \textbf{New Encoding Strategy}: GA-KAN encodes neuron connections, grid values, and depth of the KANs into the chromosomes of individuals. It searches for the optimal number of layers and connections in the KAN architecture while simultaneously optimizing the network's grid values.
    \item \textbf{New Decoding Process}: During decoding, the combination of a \textit{degradation mechanism} and zero masks allows for more efficient exploration of KAN structures across various depths, thereby facilitating a more diverse and flexible search process.
    \item \textbf{Improved Accuracy, Interpretability, and Parameter Reduction}: GA-KAN has been validated through several classification tasks, demonstrating superior accuracy to both traditional machine learning models and the standard KAN proposed in the original paper. Furthermore, the optimized KAN model improves interpretability and reduces the number of parameters.
\end{enumerate}

\section{Background and Related Works}

\subsection{Genetic Algorithm (GA)}

Among optimization methods, Evolutionary Algorithms (EAs), including Genetic Algorithms (GAs) \cite{Multi-objective-evolutionary-federated-learning, zhu2021real}, Genetic Programming (GP) \cite{2021A_genetic_programming_approach}, Evolutionary Programming (EP) \cite{1999Evolutionary_programming}, Differential Evolution \cite{2020Differential_Evolution}, and Evolution Strategies (ES) \cite{2017Evolution_strategies}, are notable for their biological inspiration, leveraging concepts like natural selection, crossover, and mutation to explore optimal solutions. Among these, GAs are considered one of the most established and widely recognized approaches \cite{tilahun2013comparison}. 

GAs mimic natural selection, or ``survival of the fittest," by combining the best genes from the fittest individuals of previous generations, as described in \textbf{Algorithm}~\ref{alg:genetic_algorithm}. They operate on a population of individuals, encoded as chromosomes. Evolutionary operations like selection, crossover, and mutation generate new individuals. Selection chooses parents based on defined criteria, while crossover swaps chromosome sections between two parents, and mutation introduces random variations. Each individual is evaluated using a fitness function, which provides a numerical value to guide the selection process \cite{michalewicz1996evolutionary}. Through successive generations, the individual with the highest fitness score becomes the optimal solution.


\begin{algorithm}[!t]
\caption{Genetic Algorithm (GA)}
\label{alg:genetic_algorithm}
\begin{algorithmic}[1]
\State \textbf{Input:} Population size $N$, Maximum generations $G$, Crossover rate $r_c$, Mutation rate $r_m$
\State \textbf{Output:} Optimal solution $S^*$
\State \textbf{Initialization:} Generate an initial population $P$ of $N$ individuals (denoted as $x$)
\State \textbf{Fitness Evaluation:} Compute fitness $f(x)$ for each individual in the population
\For{$g \gets 1$ to $G$}
    \State \textbf{Crossover:} For each pair of parents, with probability $r_c$, perform crossover to generate offspring
    \State \textbf{Mutation:} For each offspring, with probability $r_m$, randomly mutate genes
    \State \textbf{Fitness Evaluation:} Compute fitness $f(x)$ for each individual in the population
    \State \textbf{Selection:} Select parents based on fitness $f(x)$
\EndFor
\State \textbf{Return:} Individual $S^*$ with the highest fitness $f(S^*)$
\end{algorithmic}
\end{algorithm}

GA is considered one of the most popular metaheuristic algorithms in practical optimization, design, and application domains \cite{albadr2019spoken}. GAs feature relatively low computational resource requirements and more persuasive optimization strategies. Moreover, GA-based Neural Architecture Search (NAS) has gathered significant attention in the literature, showing promising performance \cite{feng2024enhancing}. For the above reasons, GA is selected for this task.

\subsection{Kolmogorov–Arnold Networks (KAN)}

In 2024, Liu \textit{et al.}\cite{KAN} made a significant contribution by introducing KANs in a pioneering paper published on arXiv, which quickly attracted widespread interest within the research community. KANs offer a compelling alternative to conventional MLPs, opening up new avenues for the development of modern deep learning models that differ from MLP architectures.

KANs are fundamentally based on the Kolmogorov-Arnold representation theorem. In 1957, Andrey Kolmogorov addressed Hilbert’s 13th problem \cite{hilbert2019mathematical} by proving that any multivariate continuous function can be represented as a sum of single-variable functions and their combinations. This theorem \cite{kolmogorov1957representation, braun2009constructive} states that any continuous multivariate function $f$ defined on a bounded domain can be expressed as a finite composition of continuous single-variable functions and their sums. For a set of variables $\mathbf{x} = x_1, x_2, . . . , x_n$, where $n$ denotes the number of variables, the multivariate continuous function $f(\mathbf{x})$ can be represented as follows:
\begin{equation}\label{eqn-kan}
f(\mathbf{x}) = f(x_1,...,x_n)=\sum_{q=1}^{2n+1} \Phi_q( \sum_{p=1}^{n} \phi_{q,p}(x_p)),
\end{equation}
here, $\phi_{q,p}:[0,1]\to \mathbb{R}$ and $\Phi_q:\mathbb{R}\to \mathbb{R}$. 

Although the Kolmogorov-Arnold representation theorem is theoretically robust, its application in neural networks remains debated. Girosi and Poggio \cite{girosi1989representation} highlighted potential challenges, such as the non-smoothness of the function $\phi_{q,p}$ in Eq.~\eqref{eqn-kan} \cite{vitushkin1954hilbert}, which could hinder practical learning. This has led to skepticism regarding its real-world utility \cite{poggio2020theoretical}.

In contrast, Liu \textit{et al.} \cite{KAN} proposed KANs, which relax the constraints of Eq.~\eqref{eqn-kan}, allowing for arbitrary width and depth instead of restricting the architecture to two layers and a fixed number of terms $(2n+1)$.

A typical KAN consists of $L$ layers, with the output for input $\mathbf{x}$ expressed as:
\begin{equation}\label{eqn-kan-2}
\mathrm{KAN}(\mathbf{x}) = (\Phi_{L-1} \circ \Phi_{L-2} \circ \cdots \circ \Phi_1 \circ \Phi_0)\mathbf{x},
\end{equation}
where $\Phi_l$ represents the function matrix of the $l^{th}$ layer. The activation function $\phi_{l,j,i}$ connects $(l,i)$ neuron in layer $l$ to $(l+1,j)$ neuron in layer $l+1$:
$$\phi_{l,j,i},\quad l=0,\cdots,L-1, \quad i=1,\cdots,n_l, \quad j=1,\cdots ,n_{l+1}.$$
The activate values between layer $l$ and $l+1$ are given in matrix form as:
$$\mathbf{x}_{l+1}=\underbrace{
\begin{pmatrix}
 \phi_{l,1,1}(\cdot) & \phi_{l,1,2}(\cdot) & \cdots  & \phi_{l,1,n_l}(\cdot) \\
 \phi_{l,2,1}(\cdot) & \phi_{l,2,2}(\cdot) & \cdots  & \phi_{l,2,n_l}(\cdot) \\
 \vdots  & \vdots &  & \vdots \\
 \phi_{l,n_{l+1},1}(\cdot) & \phi_{l,n_{l+1},2}(\cdot) & \cdots  & \phi_{l,n_{l+1},n_l}(\cdot) \\
\end{pmatrix}  
}_{\Phi_l}
\mathbf{x}_l$$

Each activation function $\phi(x)$ is defined as:
\begin{equation}
\phi (x) = w_bb(x) + w_s\mathrm{spline}(x),
\end{equation}
where $w_b$ and $w_s$ are trainable weights, $b(x)$ is a basis function (similar to residual connections), and $\mathrm{spline}(x)$ is a linear combination of B-splines:
\begin{equation}
\mathrm{spline}(x) = \sum_{i=0}^{G+k-1}c_iB_{i,k}(x),
\end{equation}
with $c_i$ as trainable parameters and $B_{i,k}(x)$ as B-spline basis functions of degree $k$ on $G$ grid intervals. The grid $G$ influences the curve's resolution and smoothness. In this work, we fix $k = 3$ and encode $G$ into a vector for evolutionary optimization.

For a specific task, the process of finding more interpretable solutions in KAN often involves pruning method \cite{KAN} that requires setting certain parameters, which may vary across different datasets. The pruning method is labor-intensive and typically requires parameter tuning, presenting a significant challenge. To address the issue, a GA-based neural architecture search can be applied to KAN. Our approach automates the search for the optimal KAN structure, reducing the need for manual design and enhancing KAN's performance in both accuracy and interpretability.

\subsection{Related Work in NAS}

Neural Architecture Search (NAS) automates the design of neural network architectures by exploring a defined search space to find optimal architectures that minimize performance metrics like validation error \cite{liu2022survey}. Early works such as NASRL \cite{NASRL} and MetaQNN \cite{MetaQNN} leveraged reinforcement learning (RL) as the optimization strategy.
Besides RL, Evolutionary Computation (EC) \cite{Introduction_to_evolutionary_computing} based NAS algorithms (ENAS) form another category in NAS.

ENAS algorithms address architecture search challenges using techniques like Evolutionary Algorithms (EA) and Particle Swarm Optimization (PSO). LargeEvo \cite{real2017Large-scale}, a pioneering ENAS method, applied GA to optimize CNN architecture and paved the way for numerous ENAS advancements. 
In light of the lack of a specific architecture search method for KAN, which serves as an alternative to MLP, the literature review of this paper focuses on related work regarding the architecture search for MLP.

Han \textit{et al.} \cite{han2021heuristic} utilized differential evolution to optimize the structural hyperparameters of MLPs, such as the number of layers and neurons, for genomic prediction.
Azad \textit{et al.} \cite{talatian2022intelligent} employed GA and PSO to optimize MLP parameters, including optimal features, hidden layers, hidden nodes, and weights, for breast cancer diagnosis.
Sarkar \textit{et al.} \cite{sarkar2022hyperparameters} employed PSO to optimize hyperparameters, including the number of hidden layers, neurons per layer, activation functions, and training functions, for solving inverse electromagnetic problems.
Martínez-Comesaña \textit{et al.} \cite{martinez2021use} applied NSGA-II to balance error and complexity in optimizing MLP architectures. 
Ansari \textit{et al.} \cite{ansari2022automatic} leveraged GA to tune parameters such as topology and learning rate.  
Mohan \textit{et al.} \cite{mohan2023novel} developed a GA-tuned superlearner for improving predictive accuracy. 
El-Hassani \textit{et al.} \cite{el2024new} proposed an RCGA framework for optimizing both the architecture and hyperparameters of MLPs.
While these methods have shown promising results in MLP optimization, it is worth noting that they are built on the assumption of full connectivity and do not explicitly account for the connections between neurons.

The proposed GA-KAN, in contrast, employs GA to explore optimal KAN architectures, enabling the discovery of sparse connection patterns beyond the fully connected assumption typically used in MLP architecture search. 

\section{The Proposed Method}

In this section, the overall framework of the proposed method, named GA-KAN, for automatically optimizing the KAN architecture is presented. 
Section~\ref{sec:Overall_Framework} provides a comprehensive overview of the framework. 
Each component of the framework will then be elaborated in detail.
The encoding strategy is presented in Section~\ref{sec:Encoding_Strategy}, while Section~\ref{sec:Decoding} focuses on the decoding method. The crossover and mutation operators are described in Section~\ref{sec:crossover_mutation}, followed by an explanation of the fitness evaluation process in Section~\ref{sec:fitnee_evaluation}.
Finally, the interpretability offered by the proposed method is illustrated in Section~\ref{sec:interpretability}.


\subsection{Overall Framework} \label{sec:Overall_Framework}

\begin{figure*}[!t]
\centering
\includegraphics[width=6in]{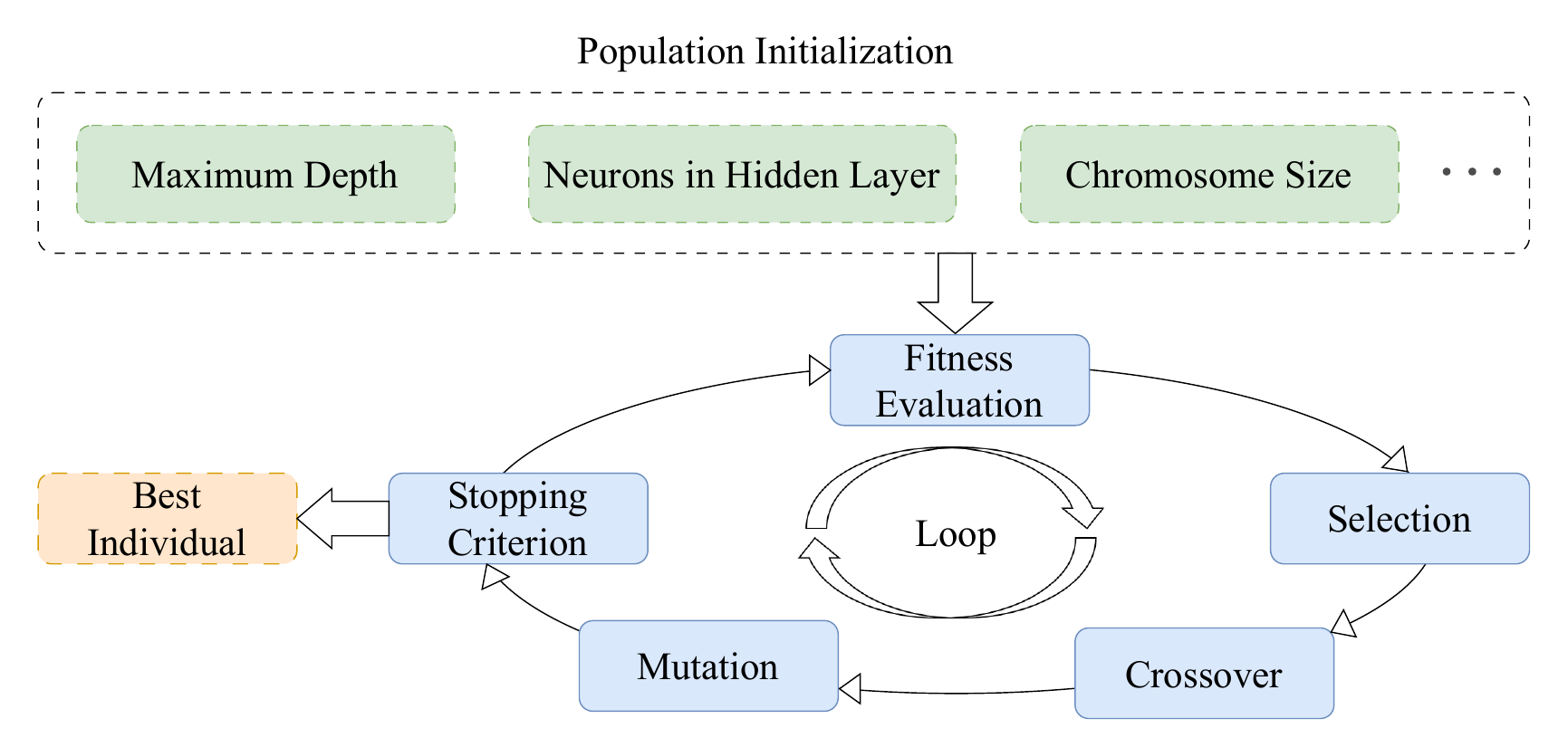}
\caption{The overall framework of the proposed method, where $t$ denotes the generation, $P_{t}$ represents the parent population, and $Q_{t}$ represents the offspring population. The next parent population, $P_{t+1}$, is selected based on a combination of $P_{t}$ and $Q_{t}$. The dashed box indicates an element, while the solid box highlights a key step in GA.}
\label{fig_1}
\end{figure*}

The overall framework of the proposed method is shown in Fig.~\ref{fig_1}. First, the search space is defined by specifying key hyperparameters, such as the maximal depth, and the maximal number of neurons in each hidden layer. These parameters collectively determine both the range of network architectures explored by the GA method and the chromosome size, which is fixed throughout the optimization process. The decoding process includes the network’s depth, with a \textit{degradation mechanism} introduced to allow decoding the chromosome to networks of varying depths. The \textit{degradation mechanism} enables the network to reduce from the maximal depth to a smaller depth (see Section~\ref{sec:Decoding} for more details). Additionally, other hyperparameters need to be defined, such as the population size and the number of generations. Next, the initial population ($P_t$) is generated with a predefined population size, where each chromosome is randomly initialized with binary values (0 or 1). 
Each individual in the population undergoes a fitness evaluation.

The algorithm then enters the GA loop, where the population undergoes selection, crossover, and mutation, followed by fitness evaluation. This iterative process continues until the stopping criterion is met. The crossover and mutation processes are discussed in Section~\ref{sec:crossover_mutation}, while fitness evaluation is detailed in Section~\ref{sec:fitnee_evaluation}.

Finally, the best-performing individual is identified as the optimal solution, representing the best KAN architecture. The optimized structure offers valuable interpretability in three main aspects: feature importance, feature selection, and symbolic formulae derived from the KAN model (see Section~\ref{sec:interpretability} for further details).

\subsection{Encoding Strategy} \label{sec:Encoding_Strategy}

\begin{figure}[!t]
\centering
\includegraphics[width=3.2in]{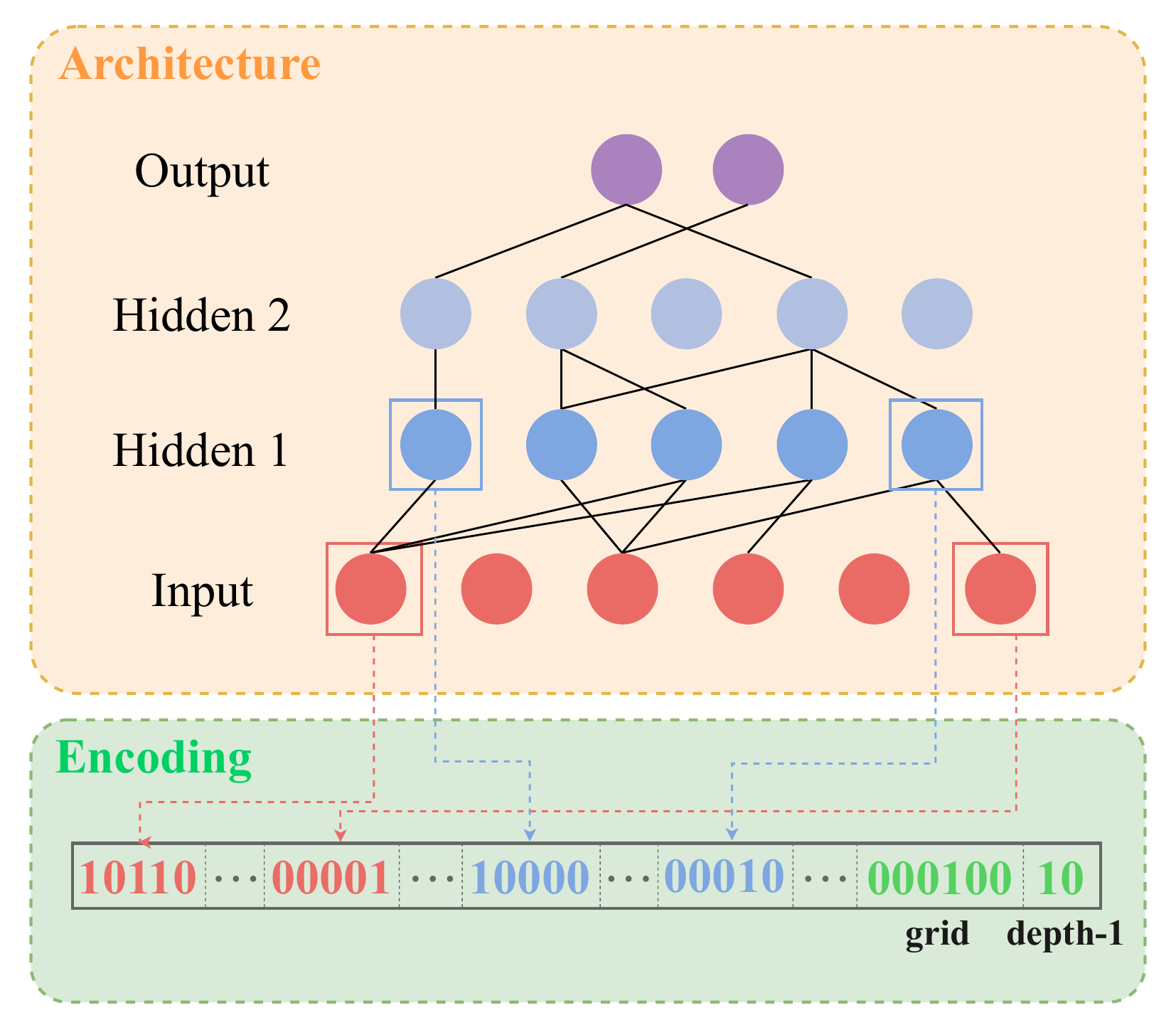}
\caption{An example of the encoded chromosome. The connections between lower-layer and upper-layer neurons are encoded using 0s and 1s, where 0 indicates no connection and 1 indicates a connection. The encoding of all neurons is organized into a chromosome. Additionally, 6 fixed bits are used to encode the grid value, and 2 fixed bits are used to encode the network depth.}
\label{fig_2}
\end{figure}

Before initiating the encoding process, the first critical step is to define the search space, which establishes the boundaries for the GA's search. The search space is defined by key hyperparameters: the number of input neurons $n$, the number of output neurons $m$, the maximal depth $d$, and the maximal number of neurons in each hidden layer $u$. While $n$ and $m$ are typically determined by the structure of the dataset, $d$ and $u$ are carefully selected to balance network complexity with available computational resources. These choices directly influence the structure of the chromosomes in GA, particularly their length. Once the search space is defined, the network with the largest possible structure within this space is determined, and the encoding of individual networks is carried out accordingly, which remains fixed throughout the optimization process.

As illustrated in Fig.~\ref{fig_2}, for a given network architecture, the encoding process captures the presence or absence of connections between neurons across layers. Specifically, in each layer $i$, where the number of input neurons is $n_{i}$ and the number of output neurons is $m_{i}$, a connection between an input neuron and an output neuron is represented by a bit value of 1, while the absence of a connection is represented by 0. The total number of encoded bits for layer $i$ is thus $n_{i} \times m_{i}$, reflecting all possible connections between neurons in adjacent layers. 

In addition to the layer encoding, each individual's chromosome also includes $b_{\mathrm{grid}}$ bits, representing the grid value of the KAN. The grid value is a critical parameter, set by default to 6 bits, allowing for a representation of values ranging from 1 to 64. This value influences the network's structural configuration, adding another layer of variability to the encoding scheme.
Furthermore, the encoding process incorporates $b_{\mathrm{depth}}$ (set by default to 2) additional bits to represent the network depth. These depth bits allow the network to decode into structures with varying depths. 
This approach ensures that the GA method has the flexibility to search through a broader range of network architectures adaptively. Since the minimal depth is 1, the 2 bits representing the depth encode the value of $\mathrm{depth}-1$, as illustrated in Fig.~\ref{fig_2}. With 2 bits, the depth can be encoded to represent depths ranging from 1 to 4. 

The total number of bits required to encode an arbitrary network architecture is given by Eq.~\eqref{eqn-total-bits}.
\begin{equation}\label{eqn-total-bits}
b_{\mathrm{total}} = \sum_{i=1}^{d} (n_{i}\times m_{i}) + b_{\mathrm{grid}} + b_{\mathrm{depth}}
\end{equation}

For instance, consider the network architecture depicted in Fig.~\ref{fig_2}: the input layer consists of 6 neurons, there are 2 hidden layers each containing 5 neurons, and the output layer has 2 neurons. The encoding of this network requires a total of 73 bits, calculated as $6\times5 + 5\times5 + 5\times2 + 6 + 2 = 73$. Therefore, each individual's chromosome in the population is composed of 73 bits.


\subsection{Decoding Chromosomes to KANs} \label{sec:Decoding}

\begin{figure}[t]
\centering
\includegraphics[width=\linewidth]{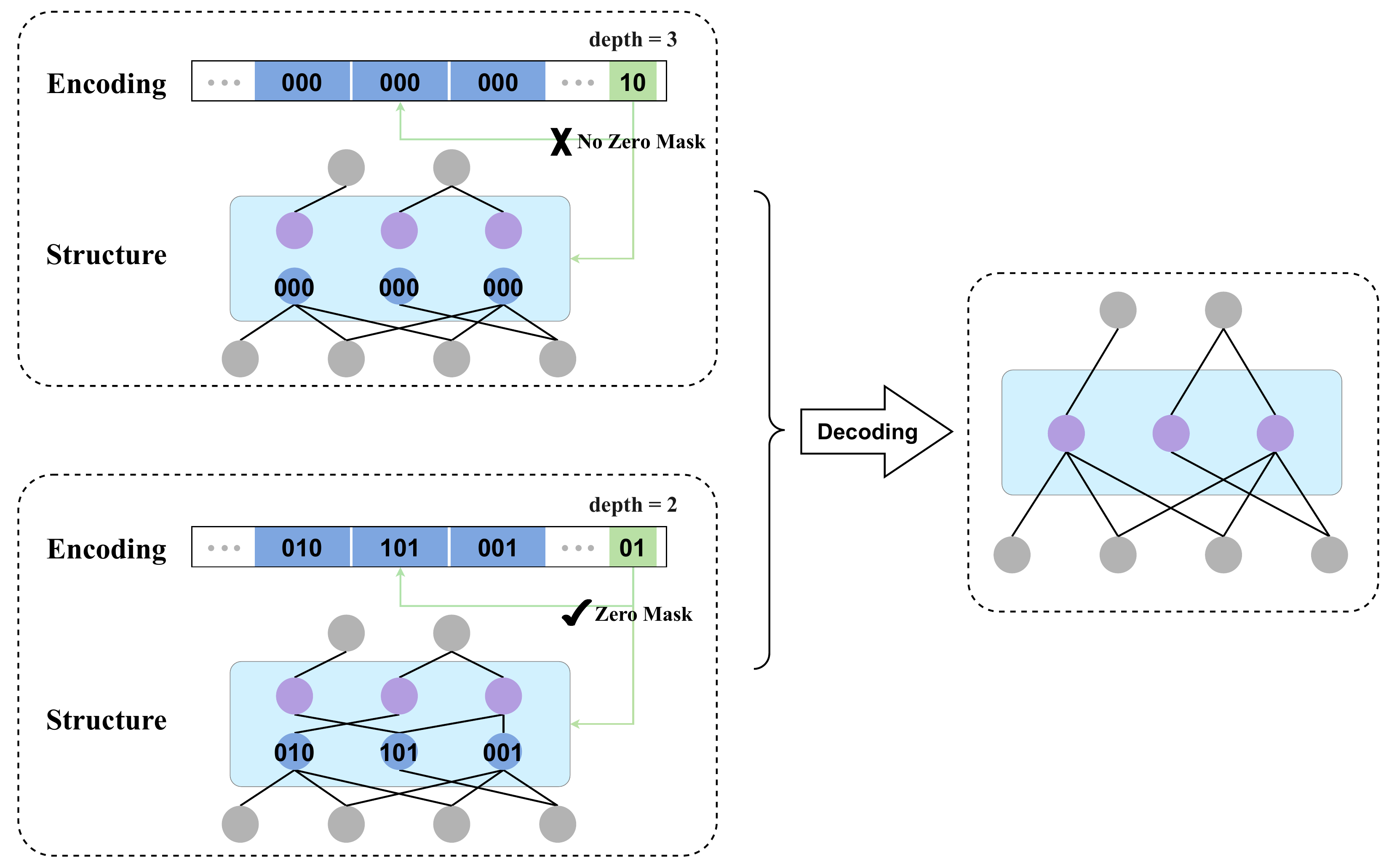}
\caption{An example of the \textit{degradation mechanism}---A process where a layer is removed if all its connections are inactive, reducing its depth.}
\label{fig:degradation_mechanism}
\end{figure}

In the proposed encoding strategy (see Section~\ref{sec:Encoding_Strategy}), each individual's chromosome, with its fixed length, defines a neural network with the maximal depth by the encoding. However, constructing networks with reduced depths is often desirable for improving computational efficiency and reducing overfitting. To enable this, the \textit{degradation mechanism} is introduced.

The \textit{degradation mechanism} is activated when an entire layer in the encoding is represented by zeros, indicating no active connections to that layer, as shown in Fig.~\ref{fig:degradation_mechanism}. Consequently, the layer is deactivated (i.e., excluded from the final network architecture), thus reducing the overall depth of the network by one. This dynamic adjustment allows for flexible network depth reduction without requiring explicit manual intervention, which is particularly advantageous in the search for efficient architectures.

As illustrated in the upper part of Fig.~\ref{fig:degradation_mechanism}, the \textit{degradation mechanism} faces a significant challenge: the probability of an entire layer being encoded as zeros is extremely low, due to the large number of bits required to represent each layer. Consequently, the GA method has a very low probability of evolving such structures. 
To enable the effective utilization of the \textit{degradation mechanism}, the network depth is directly incorporated into the encoding. 
During the encoding process, a \textit{zero mask} is proposed as a strategy to mask out an entire layer. A zero mask refers to a binary array where all the positions associated with the layer are set to 0, effectively disabling the layer's connections and causing it to be excluded from the final network structure, thus allowing the network's depth to be dynamically adjusted during decoding.
As shown in the lower part of Fig.~\ref{fig:degradation_mechanism}, a zero mask of the appropriate length is applied to the encoding of the blue layer, where all the encoding bits corresponding to that layer are set to 0. This effectively masks the specific layer, meaning that the corresponding connections are ignored during the decoding process. 

The combination of zero masks and the \textit{degradation mechanism} plays a crucial role in maintaining diversity within the population, which is essential for reducing the risk of suboptimal convergence and enhancing the GA’s ability to explore a broad range of network architectures. By applying these two methods, an individual’s chromosome can be effectively decoded into a corresponding KAN with a well-defined structure, including the number of layers, inter-layer connections, and specific grid values. The zero mask ensures that the final depth of the network matches the target depth, while the \textit{degradation mechanism} handles inactive layers. 
As a result, the generated networks vary in depth and complexity, which contributes to a more comprehensive search for optimal solutions during the evolutionary process.

\subsection{Crossover and Mutation} \label{sec:crossover_mutation}

\begin{figure}[t]
\centering
\includegraphics[width=\linewidth]{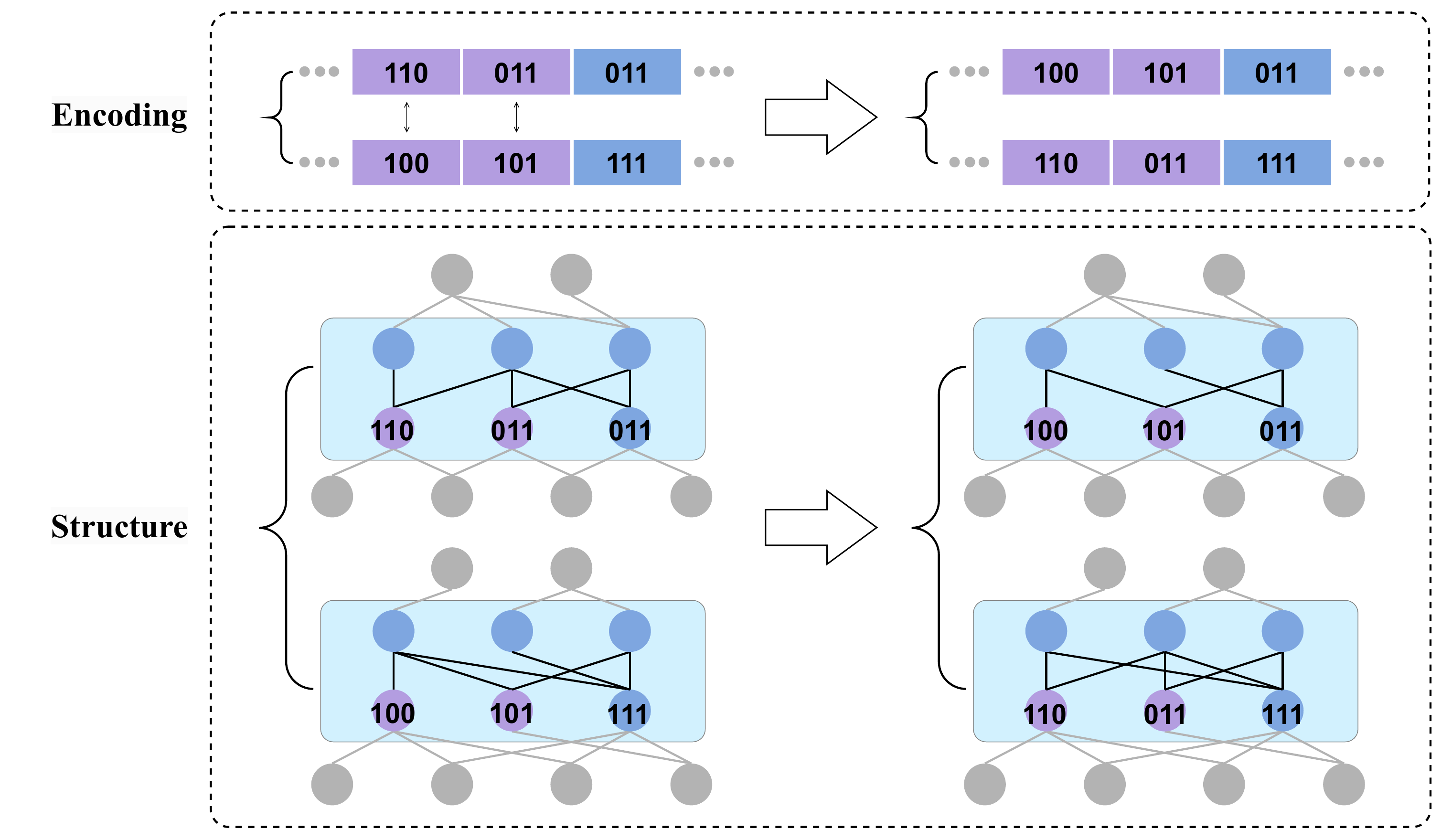}
\caption{An example of crossover. Crossover has a probability of occurring between neurons that are located in the same layer and position as those in the parent networks, indicated by the purple color.}
\label{fig:crossover}
\end{figure}

Crossover and mutation are key operations in GAs, crucial for exploring the solution space and improving population fitness. In GA-KAN, these operations are specifically designed to modify the network encoding to ensure diversity and the potential for generating high-performing offspring networks.

Crossover occurs between the same neurons of two parent networks, as illustrated in Fig.~\ref{fig:crossover}. When crossover takes place, the encoding of identical neuron positions in the parent networks is swapped based on a predetermined probability. Besides the encoding of neurons, the encoding also includes grid values and depth. For these parameters, the crossover operation employs the single-point crossover \cite{lim2017crossover}. Once all swaps are completed, two new individuals are produced. This process allows the offspring to inherit beneficial traits from both parents while introducing new combinations. This results in diverse topologies for the offspring, facilitating the exploration of a broader range of network structures in the search for optimal solutions.

Mutation affects all positions in the encoding. When a mutation occurs, binary values in the encoding are flipped---0 becomes 1, and 1 becomes 0. Although the probability of mutation is generally low to avoid excessive disruption, it plays a critical role in maintaining population diversity. Mutation introduces new connections or removes existing ones, while also affecting the depth and grid values, preventing the population from prematurely converging to local optima. In GA-KAN, mutation can introduce significant alterations in network architecture, such as forming new pathways or removing existing ones, potentially leading to improvements in network performance.

The use of both crossover and mutation in GA-KAN strikes a balance between exploration and exploitation during the search process. Crossover recombines advantageous traits from parent networks, while mutation introduces variations that can lead to novel network structures. By adjusting the probabilities of these operations, the algorithm maintains diversity in the population and continuously explores new regions of the solution space. The interplay between the two operators helps GA-KAN navigate the trade-off between refining existing solutions and discovering new, potentially better network architectures.

\subsection{Fitness Evaluation}\label{sec:fitnee_evaluation}

\textbf{Algorithm}~\ref{alg1} describes the details of the fitness evaluation function.
The fitness evaluation process is a crucial component of NAS algorithms, including GA-KAN. As described in \textbf{Algorithm}~\ref{alg1}, each individual's chromosome is decoded into a KAN structure. This structure defines not only the network’s layers and connections but also includes the specific grid values.

Once the decoding process is complete, the KAN undergoes a fitness evaluation. This begins by training the network on a designated training dataset ($\mathcal{D}_{train}$) for a fixed number of steps ($N_{steps}$), to optimize its performance based on the validation dataset ($\mathcal{D}_{val}$). The LBFGS optimizer\cite{liu1989limited} is used, following the same configuration as in the original KAN paper. Each individual’s fitness value is determined by tracking the minimum validation loss ($L_{val}$) across all training iterations.

\begin{algorithm}[!t]
\caption{Fitness Evaluation}\label{alg1}
\begin{algorithmic}[1]
\Require Individual $ind$, Training dataset $\mathcal{D}_{train}$, Validation dataset $\mathcal{D}_{val}$, Number of training steps $N_{steps} = 20$
\Ensure Fitness value $fitness$ for individual $ind$

\State Decode $ind$ into KAN structure $\mathcal{N}$
\If{$\mathcal{N}$ has invalid connections}
    \State Assign $fitness \gets +\infty$ or maximum value
\Else
    \State $min\_loss \gets +\infty$
    \For{$t = 1$ to $N_{steps}$}
        \State Train $\mathcal{N}$ on $\mathcal{D}_{train}$ using the LBFGS optimizer
        \State Compute validation loss $L_{val}$ on $\mathcal{D}_{val}$
        \If{$L_{val} < min\_loss$}
            \State $min\_loss \gets L_{val}$
        \EndIf
    \EndFor
    \State Set $fitness \gets min\_loss$
\EndIf
\State \textbf{Return} fitness value $fitness$
\end{algorithmic}
\end{algorithm}

The fitness evaluation serves a dual purpose: it not only assesses the performance of an individual but also functions as a filtering mechanism for invalid network structures. Since chromosomes are initialized randomly, decoded networks may contain invalid connections, where no viable path exists between input and output layers. In such instances, assigning a fitness value of $+\infty$ (or the maximum fitness) ensures that these non-functional networks are effectively excluded from selection in future generations. This filtering process enables the evolutionary search to concentrate on valid architectures, thereby enhancing the algorithm's efficiency and accelerating convergence toward optimal solutions.

Additionally, the fitness evaluation phase in GA-KAN ensures that promising individuals are retained by using the minimum validation loss over multiple training iterations, even if they exhibit variability in performance during the early stages of training. This approach enhances GA-KAN's ability to explore a broad range of network structures while maintaining computational feasibility.


\subsection{Interpretability}\label{sec:interpretability}

The interpretability of GA-KAN primarily stems from the inherent characteristics of KAN. Our method emphasizes three key aspects: feature importance, feature selection, and symbolic formulae extracted from the network.

For feature importance, the \texttt{feature\_score} \cite{KAN} method is used to compute the importance scores of the features through a layer-wise backpropagation process. Starting from the output layer, the method calculates scores for nodes, and edges in each layer, propagating them towards the input layer. The final scores are obtained by averaging the values across layers, indicating the contribution of each feature to the model’s output and highlighting their relative importance.

In terms of feature selection, GA-KAN encodes both network nodes and the connections between them, including features as input nodes. Throughout the evolutionary process, the encoding is optimized, allowing feature combinations that align more closely with the data distribution to achieve higher fitness scores during evaluation. This process effectively guides the model in selecting the most relevant features for the classification task.

The extraction of symbolic formulae consists of two fundamental stages. 
In the first stage, referred to as \texttt{auto\_symbolic} \cite{KAN}, the method automatically evaluates all candidate formulae by calculating their coefficients of determination---a statistical measure that indicates how well the data fit a given model, with higher values reflecting a better fit. The formula with the highest score is then selected as the connection function for each edge. This automated approach typically generates symbolic formulae of relatively high quality.
However, a manual refinement, which is entirely optional, could be employed to further improve the explainability of the symbolic formulae as the second stage. In this stage, visual analysis is used to examine the shapes of the connection functions, allowing formulae with similar structures to be fixed and assigned as the connection functions. After finalizing the connection functions, the symbolic formulae are further refined through training until they converge to their optimal representation. This process not only enhances the model's interpretability but also provides explicit symbolic relationships that illustrate how predictions are generated.


\section{Experiment Design}\label{sec:experiment_design}

\begin{table*}[!t]
\caption{Datasets from the UCI Repository}
\centering
\begin{tabular*}{\textwidth}{@{\extracolsep{\fill}} c l c c c c c} 
\toprule
\textbf{No.} & \textbf{Dataset} & \textbf{\# Classes} & \textbf{\# Instances} & \textbf{Instances per Class} & \textbf{\# Features} & \textbf{Missing Values} \\ \midrule
1 & Iris\cite{iris_53}      & 3 & 150   & 50:50:50  & 4     & No \\
2 & Wine\cite{wine_109}     & 3 & 178   & 59:71:48  & 13    & No \\ 
3 & Rice\cite{rice_545}     & 2 & 3810  & 2180:1630 & 7     & No \\
4 & WDBC\cite{wdbc_17}      & 2 & 569   & 357:212   & 30    & No \\
5 & Raisin\cite{raisin_850} & 2 & 900   & 450:450   & 7     & No \\
\bottomrule
\end{tabular*}
\label{tab:uci_datasets}
\end{table*}

In this section, the experimental setup used to evaluate the performance of the proposed GA-KAN is outlined. The evaluation was conducted across various benchmark datasets, and the configuration of parameters was carefully selected to ensure a balanced trade-off between exploration and exploitation in GA. GA-KAN is benchmarked against peer competitors to demonstrate its unique performance. Additionally, given that KAN is undergoing rapid development with frequent new releases, stability in the experiments was ensured by using version 0.2.1 throughout this study.

\subsection{Benchmark Datasets} \label{sec:benchmark_datasets}


To validate that GA-KAN optimizes the network structure of KAN without requiring manual tuning of pruning parameters, two simple toy datasets, used in the original KAN paper, were selected to demonstrate both the autonomy and effectiveness of GA-KAN. The two mathematical formulas used in the experiments are presented in Eq.~\eqref{eqn-1}, which were employed to construct the corresponding datasets.

\begin{subequations}\label{eqn-1}
  \begin{align}
    f(x, y) & = e^{\sin(\pi x) + y^2} \\
    f(x, y) & = xy
  \end{align}
\end{subequations}

As a pioneering study, the original KAN paper \cite{KAN} did not conduct an in-depth exploration of classification tasks, primarily focusing on simpler, lower-dimensional datasets. 
Additionally, it highlighted that the major bottleneck of KANs is their slow training process, with KANs being typically 10x slower than MLPs, even with the same number of parameters.
This study employs a GA for NAS, where training KANs is part of the fitness evaluation. As the evaluation is often the most time-consuming stage in ENAS algorithms \cite{liu2021survey-ENAS}, the computational cost of the proposed GA-KAN could be much higher than other ENAS of evolving MLPs. 
Therefore, for classification tasks, this study selects relatively low-dimensional datasets for validation. Additionally, these datasets exhibit relatively balanced class distributions without significant data imbalance or missing values, making them easier to work with and reducing the need for complex data preprocessing. This ensures a fair horizontal comparison, allowing for a clear positioning of GA-KAN to other methods.

For these purposes, this study employs five publicly available datasets from the UCI Machine Learning Repository: Iris, Wine, Raisin, Rice, and WDBC. Comprehensive details on these datasets are provided in \autoref{tab:uci_datasets}. These datasets cover various classification tasks with different numbers of instances, features, and class distributions. 
This assortment of datasets provides diverse dimensions and distributions, facilitating a comprehensive evaluation of GA-KAN.

\subsection{Parameter Settings} \label{sec:parameter_settings}

In evolutionary algorithms, the performance is highly sensitive to parameter configurations, particularly in terms of crossover, mutation rates, and population size. The parameters used in this study, listed in \autoref{tab:parameter_settings}, were chosen based on preliminary experiments and prior studies to ensure a balanced approach between exploration and exploitation in the search process. The detailed explanations of each parameter in the table are provided as follows.

\begin{table}[!t]
\caption{Parameter Settings}
\centering
\begin{tabularx}{0.48\textwidth}{|>{\centering\arraybackslash}X|c|} \hline
\textbf{Parameters}  & \textbf{Value} \\ \hline
\multicolumn{2}{|c|}{\textbf{Network Parameters}} \\ \hline
maximal depth & 4 \\ \hline
maximal number of neurons per hidden layer & 5 \\ \hline
maximal grid value & 64 \\ \hline
\multicolumn{2}{|c|}{\textbf{GA Parameters}} \\ \hline
crossover rate          & 0.9           \\ \hline
mutation rate           & 0.5           \\ \hline
population size         & 100           \\ \hline
number of generations   & 20            \\ \hline
\multicolumn{2}{|c|}{\textbf{Fitness Evaluation}} \\ \hline
optimizer               & LBFGS         \\ \hline
epochs                  & 20            \\ \hline
\end{tabularx}
\label{tab:parameter_settings}
\end{table}

1) Network Parameters: A network architecture with a maximal depth of $4$ layers, each containing $5$ neurons, was employed. This structure was chosen considering the relatively small number of features in the selected dataset. Setting a larger number of layers and neurons would result in a vast search space, affecting search efficiency. Additionally, the grid exploration range was set between 1 and 64. Therefore, these parameters were chosen to ensure a manageable search space while maintaining sufficient capacity for learning.

2) GA Parameters: The crossover rate was set to $0.9$, adhering to the widely accepted common settings in the academic community\cite{hassanat2019choosing}, to facilitate the combination of advantageous traits from parent networks. A mutation rate of $0.5$ was adopted to introduce sufficient variability and prevent premature convergence. A population size of $100$ was selected to maintain a broad search space, and the number of generations was limited to $20$, providing a balance between search depth and computational efficiency.

3) Fitness Evaluation: The LBFGS optimizer was employed for training, using full-batch gradient descent. The training was conducted for $20$ epochs. The dataset was divided into a training set, a validation set, and a test set, ensuring that the test set proportion matched that used by peer competitors on the respective dataset. Additionally, no preprocessing steps were applied to the dataset, apart from from label value conversion.

\subsection{Peer Competitors} \label{sec:peer_competitors}

A set of established peer competitors was selected to evaluate GA-KAN thoroughly. These algorithms were selected based on two criteria: performance on benchmark datasets and relevance to classification tasks. By comparing GA-KAN with established methods, this study aims to highlight its competitiveness and effectiveness in discovering optimal network architectures for classification tasks.

The peer competitors include widely adopted classification algorithms such as support vector machines (SVMs) \cite{cortes1995support, sonoda2022multiple}, random forests (RFs) \cite{breiman2001random, sun2019surrogate}, multilayer perceptrons (MLPs) \cite{rumelhart1986learning, gaikwad2019efficient}, k-nearest neighbors (KNNs) \cite{cover1967nearest, zhang2021challenges}, as well as other methods that have demonstrated strong performance on the datasets in previous studies. These methods have demonstrated robust performance across various UCI benchmark datasets and serve as reliable baselines for comparison. Since no prior studies focus on optimizing KAN architectures, this study chose these comparison algorithms as alternatives. Additionally, standard KAN settings are also used as comparison algorithms. This study selected two standard KAN configurations: one with the structure [$d, 2d+1, C$], where $d$ is the number of features in the dataset and $C$ is the number of classes, both determined by the dataset structure. Besides, $2d+1$ has been theoretically validated by the Kolmogorov-Arnold representation theorem and is a commonly referenced configuration in the KAN literature. The other configuration matches the largest network setting used in the GA-KAN approach. The grid values for both KAN configurations are determined using the grid extension method~\cite{KAN}.


\section{Results and Analysis}\label{sce:results_analysis}

This section presents a comprehensive analysis of the experimental results. First, the performance of GA-KAN in validating the symbolic formulae used in the two toy datasets from the original KAN paper is evaluated, referencing Eq.~\eqref{eqn-1}, along with a visualization of the optimal structure identified by GA-KAN. Next, the classification performance of GA-KAN on the benchmark datasets, presented in \autoref{tab_results-uci_1} and \autoref{tab_results-uci_2}, is evaluated through comparison with peer competitors to highlight its effectiveness in classification tasks. The convergence behavior of GA-KAN is visualized and analyzed to further demonstrate its capability and efficiency. Finally, the interpretability of the network on two simple datasets is illustrated, focusing on feature importance, feature selection, and symbolic formulae extracted from the network.

\subsection{Toy Datasets Analysis}
In the original KAN paper, several formulae were provided to assess the interpretability of KAN. However, configuring the network requires setting up a larger initial network, followed by pruning operations using penalty factors (e.g., $\lambda$ and $\lambda_{\mathrm{entropy}}$ ) \cite{KAN} to find the optimal structure. Different formulae may require different parameter values, and determining the maximum network size also presents challenges. At this stage, expert knowledge is often required, which can introduce additional challenges.

\begin{figure}[htbp]
    \centering
    \subfloat[$e^{\sin(\pi x) + y^2}$]{
        \includegraphics[width=0.4\linewidth]{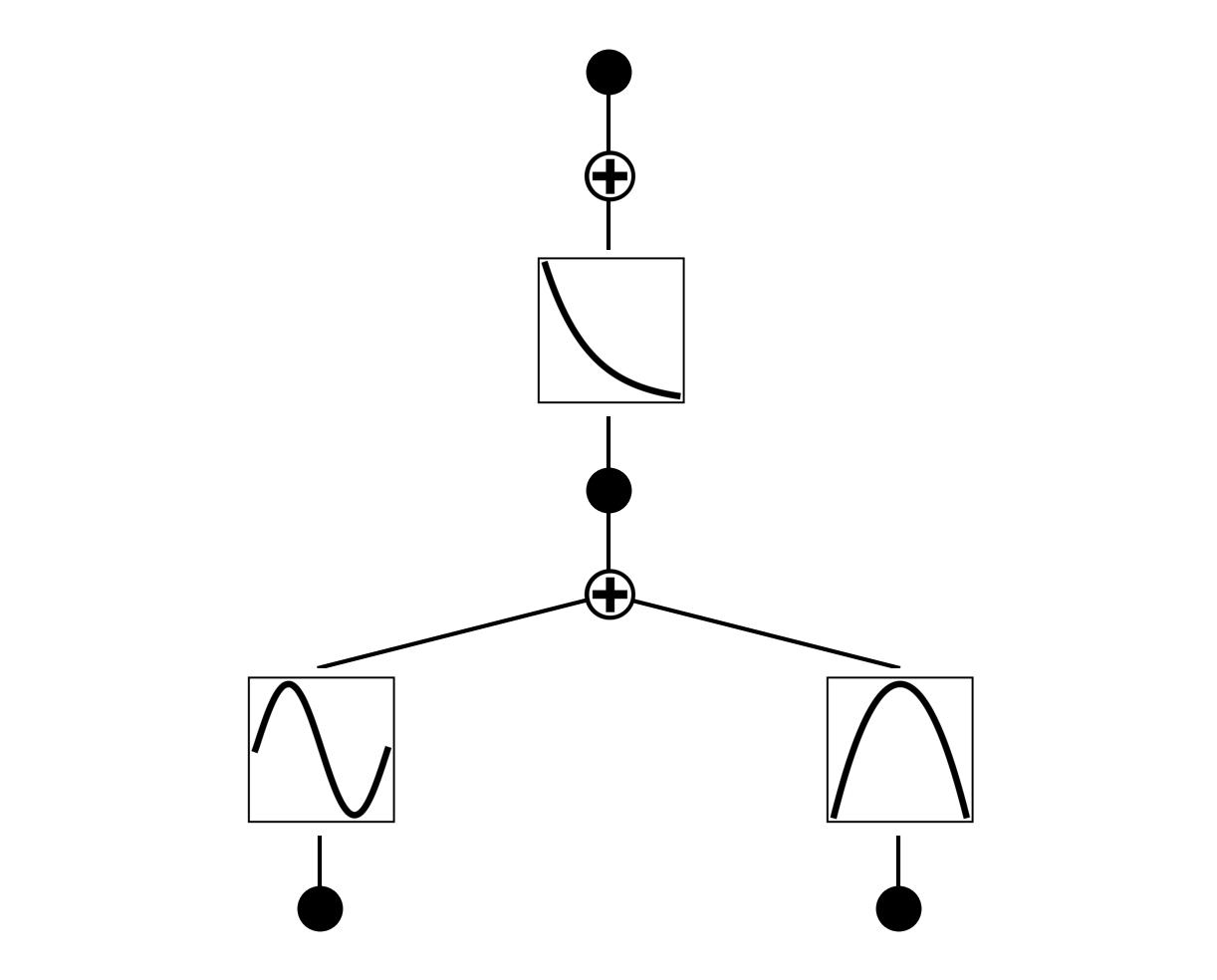}
        \label{fig:e(sin(pi_x)+y^2)}
    }
    \quad 
    \subfloat[$xy$]{
        \includegraphics[width=0.4\linewidth]{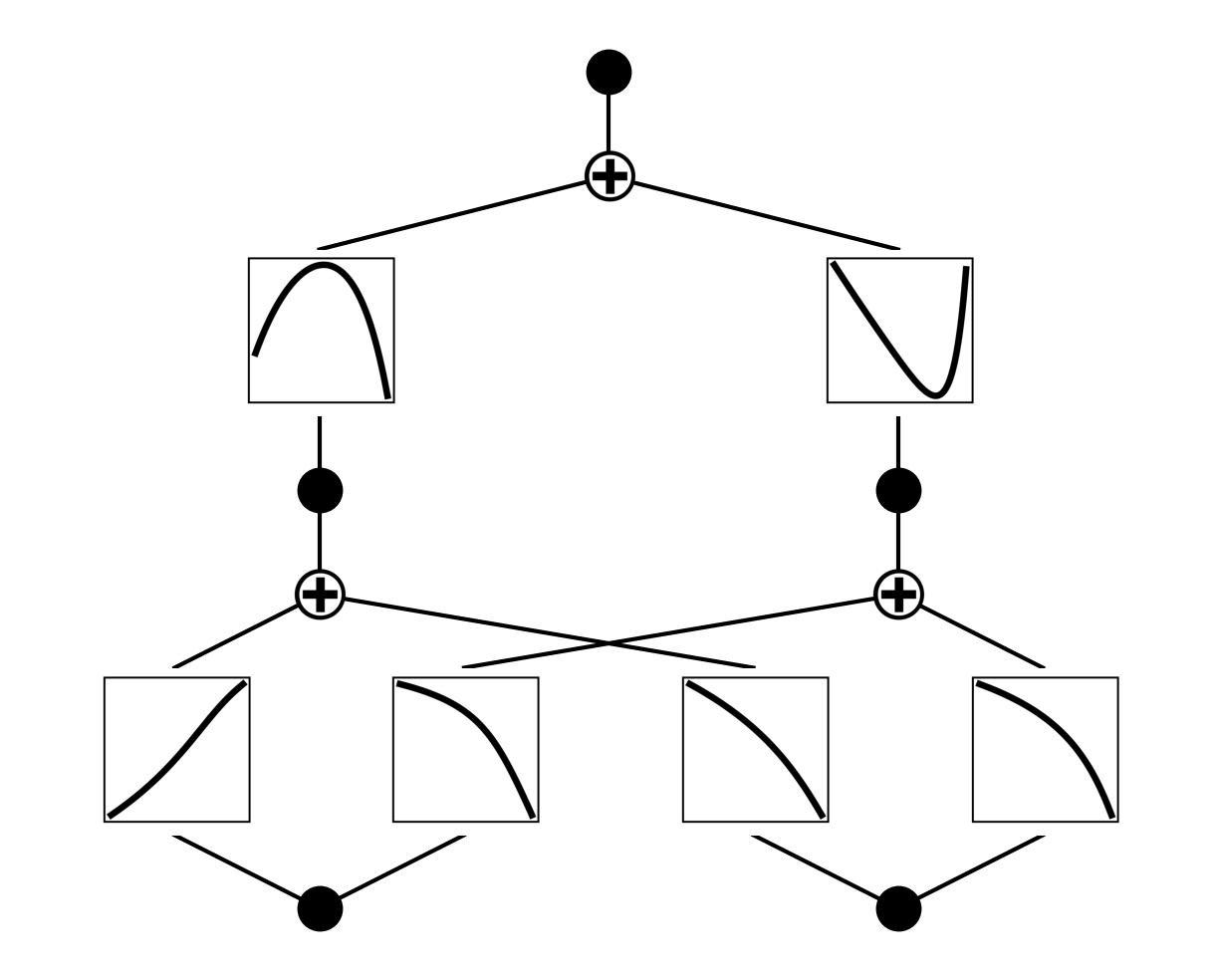}
        \label{fig:xy}
    }
    \caption{Results on the two toy datasets.}
    \label{results_formula}
\end{figure}

GA-KAN addresses these issues by automatically searching for the optimal structure without requiring parameter tuning or prior knowledge. A couple of formulae from the original paper of KAN were selected to verify the effectiveness of GA-KAN, and the results are shown in Fig.~\ref{results_formula}.

As shown in Fig.~\ref{fig:e(sin(pi_x)+y^2)}, GA-KAN successfully discovers the optimal structure for the formula $f(x, y) = e^{\sin(\pi x) + y^2}$. Fig.~\ref{fig:e(sin(pi_x)+y^2)} shows the KAN architecture in two layers. In the first layer, two distinct lines can be observed: the first line represents the sine function applied to $\pi x$, corresponding to the $\sin(\pi x)$ component of the formula, while the second line captures the square function, representing $y^2$. Moving to the second layer, an exponential curve is visible, which matches the $e^z$ part of the formula, where $z$ represents $\sin(\pi x) + y^2$. The overall structure elegantly combines these components, accurately representing $f(x, y) = e^{\sin(\pi x) + y^2}$, thus providing a clear visualization of how GA-KAN constructs and interprets this symbolic formula. The decomposition of the formula into distinct functional components in the network further highlights GA-KAN’s ability to search for and model complex mathematical expressions.

Similarly, Fig.~\ref{fig:xy} illustrates that GA-KAN has also successfully identified the optimal structure corresponding to the formula $f(x,y)=xy$. In this case, the left side of the network represents the three-line structure encoding $-(x-y)^2$, while the right side captures the structure encoding $(x+y)^2$. The combination of these two expressions leads to the difference $(x+y)^2 - (x-y)^2$, which simplifies to $4xy$. This demonstrates how GA-KAN interprets and manipulates mathematical structures, leveraging symmetry and transformation to represent the multiplication operation $f(x,y)=xy=\frac{(x+y)^2 - (x-y)^2}{4}$. Through this process, GA-KAN effectively models the interaction between variables and presents a simplified yet powerful representation of the underlying function.

\subsection{Classification Performance Evaluation}

\begin{table}[!t]
\caption{Results on Iris, Wine, and Rice Datasets}
\centering
\begin{tabular*}{0.48\textwidth}{@{\extracolsep{\fill}} c c c c} 
\toprule
\textbf{Dataset} & \textbf{Algorithm} & \textbf{Accuracy} & \textbf{AUC} \\ \midrule
\multirow{6}{*}{Iris} 
&   DT              &       98.00\%     &       -       \\
&   RF              &       99.33\%     &       -       \\
&   KNN             &       100.00\%    &       -       \\
&   KAN[4,9,3]      &       100.00\%    &       -       \\
&   KAN[4,5,5,5,3]  &       100.00\%    &       -       \\
&   \textbf{GA-KAN }& \textbf{100.00\%} &       -       \\
\midrule
\multirow{7}{*}{Wine} 
&   KAN[13,5,5,5,3] &     92.59\%       &       -       \\
&   SVM             &     93.07\%       &       -       \\ 
&   MLR             &     93.81\%       &       -       \\
&   RF              &     95.93\%       &       -       \\
&   LDA             &     98.31\%       &       -       \\
&   KAN[13,27,3]    &     100.00\%      &       -       \\
&   \textbf{GA-KAN }& \textbf{100.00\%} &       -       \\
\midrule
\multirow{7}{*}{Rice} 
&   SVM             & 92.83\%           & -             \\
&   MLP             & 92.86\%           & -             \\
&   LR              & 93.02\%           & -             \\
&   3H2D\cite{rice} & 94.09\%           & -             \\
&   KAN[7,5,5,5,2]  & 94.23\%           & 0.983         \\
&   KAN[7,15,2]     & 94.23\%           & 0.984         \\
& \textbf{GA-KAN}   & \textbf{95.14\%}  & \textbf{0.985}\\
\bottomrule
\end{tabular*}
\label{tab_results-uci_1}
\end{table}

\begin{figure*}[htbp]
    \centering
    \subfloat[Iris Dataset.]{%
        \includegraphics[width=0.325\textwidth]{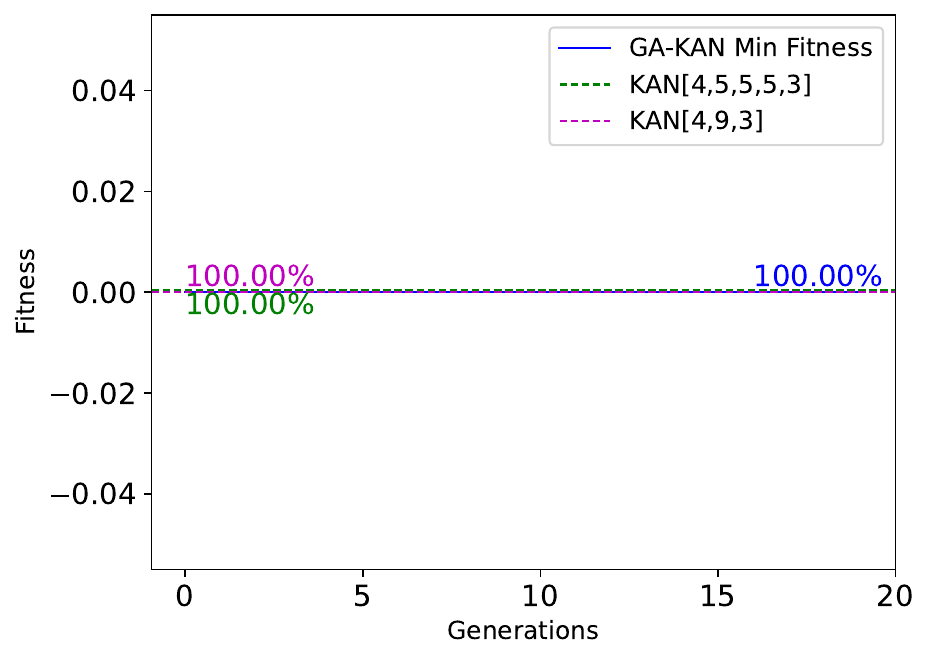}%
        \label{fig:results_iris}
    }
    \subfloat[Wine Dataset.]{%
        \includegraphics[width=0.325\textwidth]{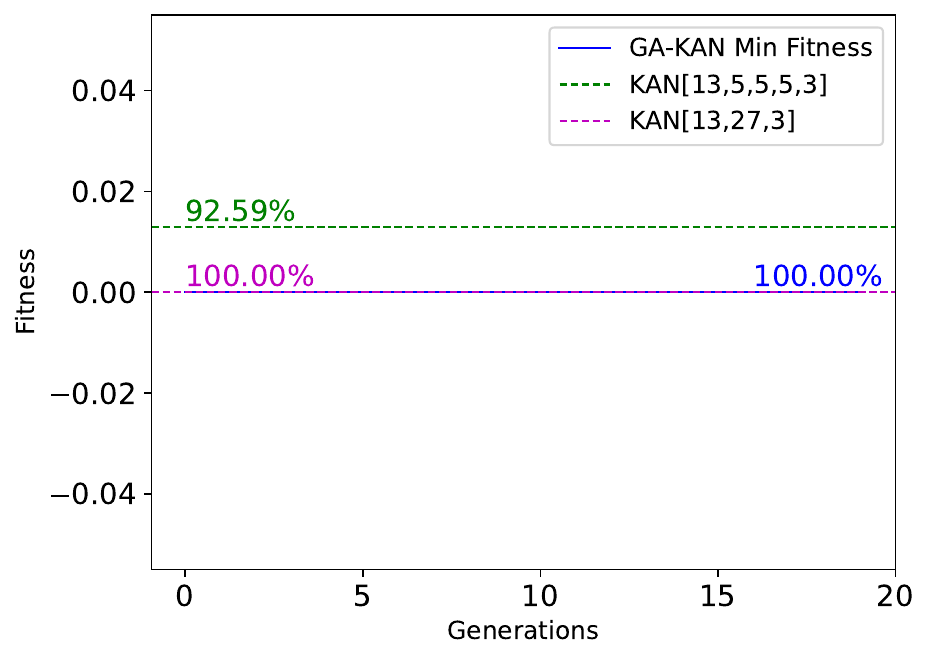}%
        \label{fig:results_wine}
    }
    \subfloat[Rice Dataset.]{%
        \includegraphics[width=0.325\textwidth]{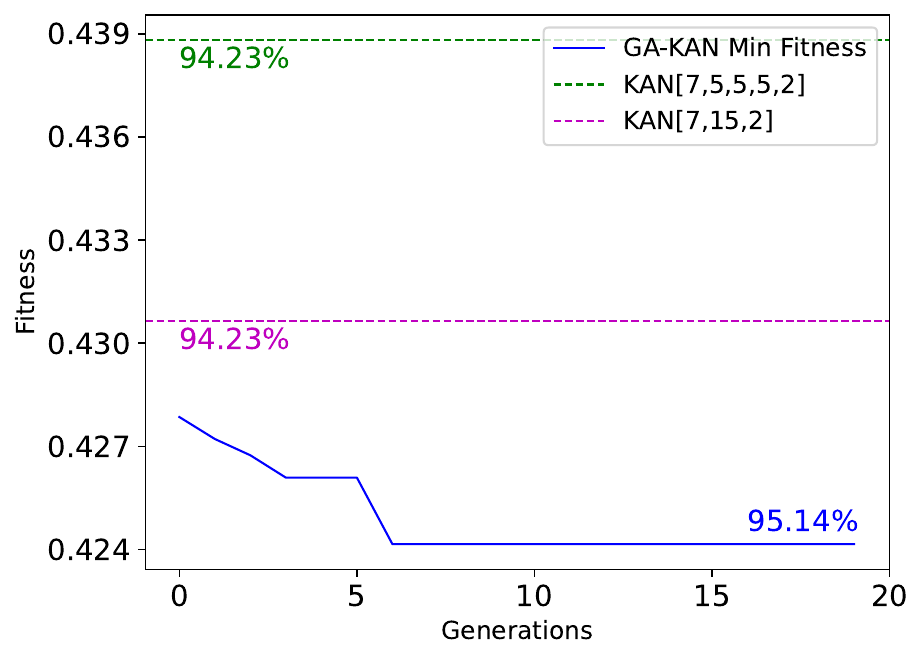}%
        \label{fig:results_rice}
    }
    
    \subfloat[WDBC Dataset.]{%
        \includegraphics[width=0.325\textwidth]{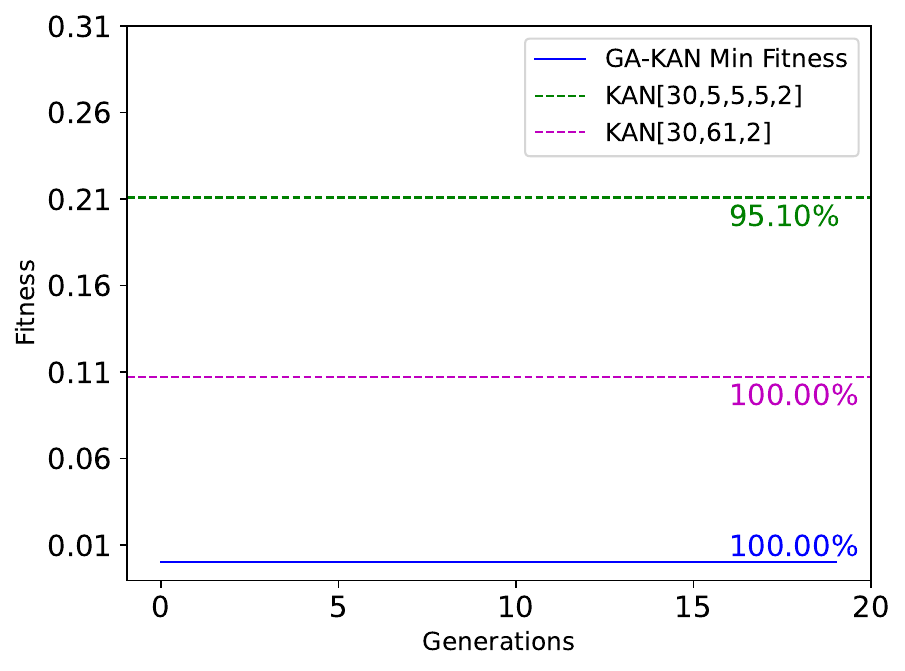}%
        \label{fig:results_wdbc}
    }
    \subfloat[Raisin Dataset.]{%
        \includegraphics[width=0.325\textwidth]{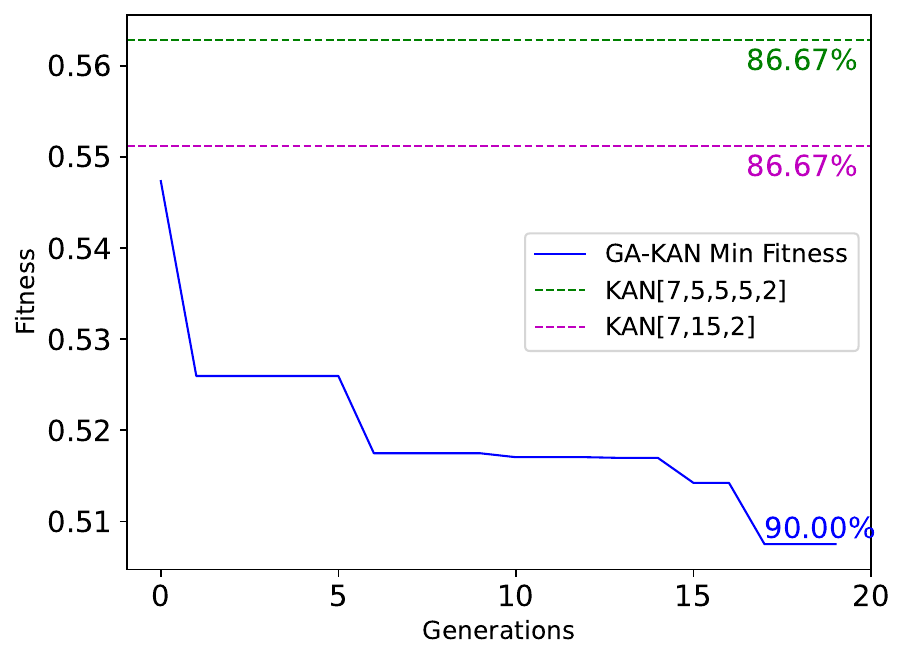}%
        \label{fig:results_raisin}
    }
    \caption{The evolution of minimum fitness values over generations for the GA-KAN optimization process across multiple datasets. The blue solid line indicates the performance of GA-KAN, while the green and purple dashed lines show the performances of the two comparative KANs. The percentage values correspond to the performance of each network on the test set.}
    \label{results_datasets}
\end{figure*}

This section presents the performance of GA-KAN on the five datasets along with a comparative analysis against its peer competitors. 
For the Iris and Wine datasets, which are three-class classification problems, the algorithms compared in the referenced studies \cite{chicho2021machine, ahammed2018predicting} did not compute AUC values. Similarly, although the Rice dataset is a binary classification problem, AUC was not reported in the referenced study \cite{rice}. Therefore, in \autoref{tab_results-uci_1}, we primarily focus on comparing accuracy.
In contrast, for the binary classification problems of the WDBC and Raisin datasets, where the compared algorithms in the referenced studies \cite{naji2021machine_WDBC, raisin} computed AUC values, we provide a comprehensive comparison of both accuracy and AUC, with the results listed in \autoref{tab_results-uci_2}.
Fig.~\ref{results_datasets} illustrates the optimization performance of GA-KAN on five different datasets, with an evaluation against standard manually designed KAN architectures. 

\begin{table}[!t]
\caption{Results on WDBC and Raisin Datasets}
\centering
\begin{tabular*}{0.48\textwidth}{@{\extracolsep{\fill}} c c c c} 
\toprule
\textbf{Dataset} & \textbf{Algorithm} & \textbf{Accuracy} & \textbf{AUC} \\ \midrule
\multirow{8}{*}{WDBC} 
&   KNN             &       93.70\%     &   0.952           \\
&   DT              &       95.10\%     &   0.945           \\
&   KAN[30,5,5,5,2] &       95.10\%     &   0.949           \\
&   LR              &       95.80\%     &   0.947           \\
&   RF              &       96.50\%     &   0.960           \\
&   SVM             &       97.20\%     &   0.966           \\
&   KAN[30,61,2]    &       100.00\%    &   1.000           \\
&   \textbf{GA-KAN} & \textbf{100.00\%}  & \textbf{1.000}    \\
\midrule
\multirow{10}{*}{Raisin} 
&   NB                      & 83.67\%           & 0.920             \\
&   KNN                     & 85.11\%           & 0.910             \\
&   DT                      & 85.11\%           & 0.866             \\
&   RF                      & 85.56\%           & 0.926             \\
&   MLP                     & 86.67\%           & 0.927             \\
&   KAN[7,15,2]             & 86.67\%           & 0.935             \\
&   KAN[7,5,5,5,2]          & 86.67\%           & 0.938             \\
&   SVM                     & 87.11\%           & 0.928             \\
&   SVM+GA \cite{raisin}    & 87.67\%           & 0.930             \\
&   \textbf{GA-KAN}         & \textbf{90.00\%}  & \textbf{0.938}    \\
\bottomrule
\end{tabular*}
\label{tab_results-uci_2}
\end{table}

\textbf{1) Iris Dataset:} 
Paper \cite{chicho2021machine} evaluated the classification performance of various algorithms on the Iris dataset. In this study, the Iris dataset was split into 80\% training, 10\% validation, and 10\% testing sets, ensuring the test set size remained consistent with \cite{chicho2021machine}. The results are summarized in \autoref{tab_results-uci_1}, where both GA-KAN and manual KAN achieved an outstanding accuracy of 100\%.
While GA-KAN achieves the same high accuracy as manual KAN, it offers the added benefit of generating multiple network structures with equivalent performance. This flexibility is particularly advantageous when factors such as the number of connections are considered. Furthermore, as shown in \autoref{tab:comparison_of_parameters}, GA-KAN achieves the same outstanding accuracy with significantly fewer parameters. 
Additionally, one solution was selected as an interpretability example, as detailed in Section~\ref{sec_Interpretability}.

\textbf{2) Wine Dataset:} 
In \cite{ahammed2018predicting}, the classification performance of various algorithms on the Wine dataset was assessed. In this study, the Wine dataset was split into 70\% training and 30\% testing sets, consistent with the setup in \cite{ahammed2018predicting}. To further refine the training process, 20\% of the training data was used for validation. The results, summarized in \autoref{tab_results-uci_1}, demonstrate that both GA-KAN and KAN achieved an outstanding accuracy of 100\%.
Although GA-KAN achieves accuracy comparable to manual KAN, it introduces the advantage of generating diverse network structures with equivalent performance. This capability provides greater adaptability in scenarios where structural considerations, such as the number of connections, are critical. Moreover, as illustrated in \autoref{tab:comparison_of_parameters}, GA-KAN achieves this excellent performance with significantly fewer parameters. To further explore its potential, a specific solution was selected to illustrate interpretability, as described in Section~\ref{sec_Interpretability}.

\textbf{3) Rice Dataset:} 
In \cite{rice}, 10 different machine learning methods were evaluated on the Rice dataset, with the top three methods listed in \autoref{tab_results-uci_1}. In this study, the Wine dataset was split into 80\% training and 20\% testing sets, consistent with the setup in \cite{rice}. To further refine the training process, 20\% of the training data was used for validation. The results are summarized in \autoref{tab_results-uci_1}. GA-KAN achieved the best accuracy of 95.14\%, and in this experiment, the AUC was also calculated. Compared to the two baseline KANs, GA-KAN achieved the highest AUC of 0.985.

\textbf{4) WDBC Dataset:} In \cite{naji2021machine_WDBC}, five machine learning algorithms were applied to the WDBC dataset, with the dataset split into 75\% for training and 25\% for testing. To ensure comparability, the same 25\% test set split was used in our experiment, with 20\% of the training set allocated for validation. Accuracy and AUC are presented in \autoref{tab_results-uci_2}, along with two baseline KAN configurations. 
As shown, GA-KAN outperformed all other methods on the WDBC dataset, achieving a perfect accuracy of 100.00\% and an AUC score of 1.0. 

\textbf{5) Raisin Dataset:} The study in \cite{raisin} evaluated various methods on the Raisin dataset, reporting accuracy and AUC results. Additionally, \cite{raisin} proposed the SVM+GA method, which combines GA for feature selection with SVM for classification, achieving an accuracy of 87.67\%. In our experiment, the Raisin dataset was divided into 80\% training, 10\% validation, and 10\% testing sets, with the test set size kept consistent with \cite{raisin} for comparability. The results are presented in \autoref{tab_results-uci_2}. For additional comparisons, two manually designed KAN architectures were included as baselines. As shown, GA-KAN outperforms all other methods, achieving the highest accuracy of 90.00\% and an AUC score of 0.938. 

\begin{table}[!t]
\caption{Parameter Count Comparison with Standard KAN Models}
\centering
\begin{tabular*}{\linewidth}{@{\extracolsep{\fill}} c l c c c} 
\toprule
\textbf{No.} & \textbf{Dataset} & \textbf{2-layer KAN} & \textbf{4-layer KAN} & \textbf{GA-KAN} \\ \midrule
1 & Iris        & 882       & 1360      & \textbf{156}       \\
2 & Wine        & 6912      & 2340      & \textbf{390}       \\ 
3 & Rice        & 1620      & 1330      & \textbf{585}      \\
4 & WDBC        & 27328     & 4410      & \textbf{1560}      \\
5 & Raisin      & 3240      & 1805      & \textbf{1064}      \\
\bottomrule
\end{tabular*}
\label{tab:comparison_of_parameters}
\end{table}

Fig.~\ref{results_datasets} illustrates the evolution of minimum fitness values over generations for GA-KAN, compared against two baseline KAN configurations. 
The results show that GA-KAN consistently achieves or surpasses the performance of the comparative KAN configurations on all datasets. Notably, it demonstrates stability and optimization capability, with significant improvements observed in datasets like Rice and Raisin. This highlights the effectiveness of GA-KAN in adapting to diverse datasets and delivering superior results.

\autoref{tab:comparison_of_parameters} compares the parameter counts of GA-KAN and the standard KAN models used in the experiments. The parameter count is calculated by summing the number of elements in all trainable parameters actively involved in the learning process. Only parameters updated during training are included, with fixed parameters excluded. This can be expressed mathematically by Eq. \ref{eqn-sum-paras}: 
\begin{equation}\label{eqn-sum-paras}
T = \sum_{p\in P}\mathrm{numel} (p),
\end{equation}
where $T$ denotes the total number of trainable parameters, $P$ represents the set of parameters actively optimized during training, and $\mathrm{numel}(p)$ indicates the total number of elements in parameter $p$.

In summary, GA-KAN not only achieves outstanding performance in terms of accuracy and AUC but also demonstrates significantly lower parameter counts, outperforming all other methods. Its consistent and competitive results highlight its robustness and effectiveness across all five datasets.

\begin{figure}[!t]
    \centering
        \includegraphics[width=0.98\linewidth]{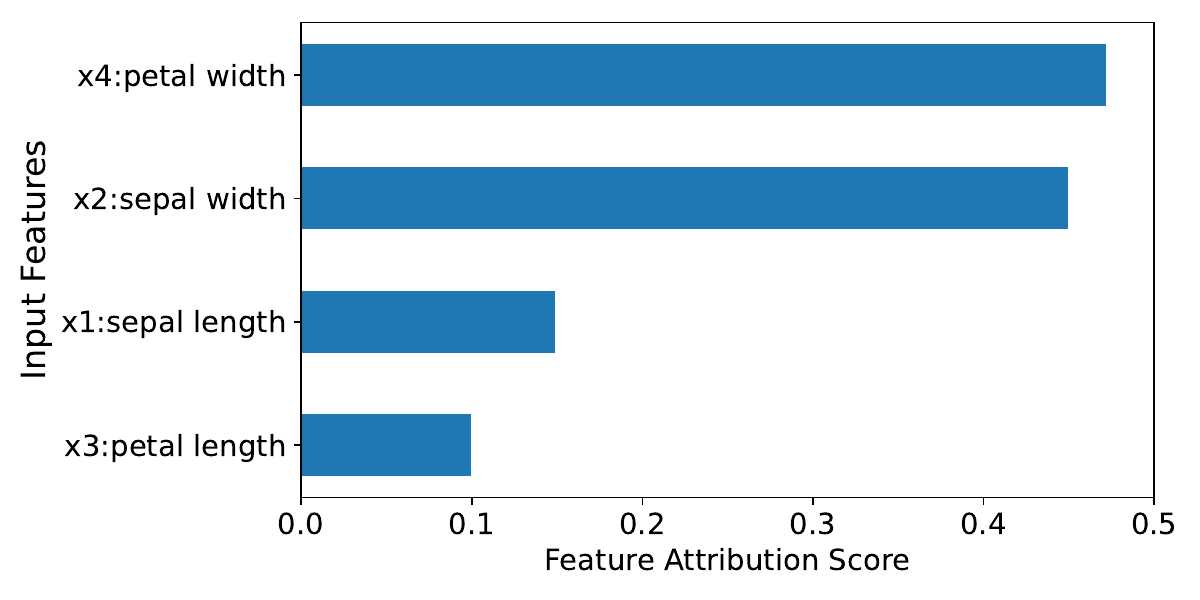}%
    \caption{Feature attribution scores on Iris dataset.}
    \label{interp:iris}
\end{figure}

\begin{figure}[!t]
    \centering
    \subfloat[The original structure.]{%
        \includegraphics[width=0.8\linewidth]{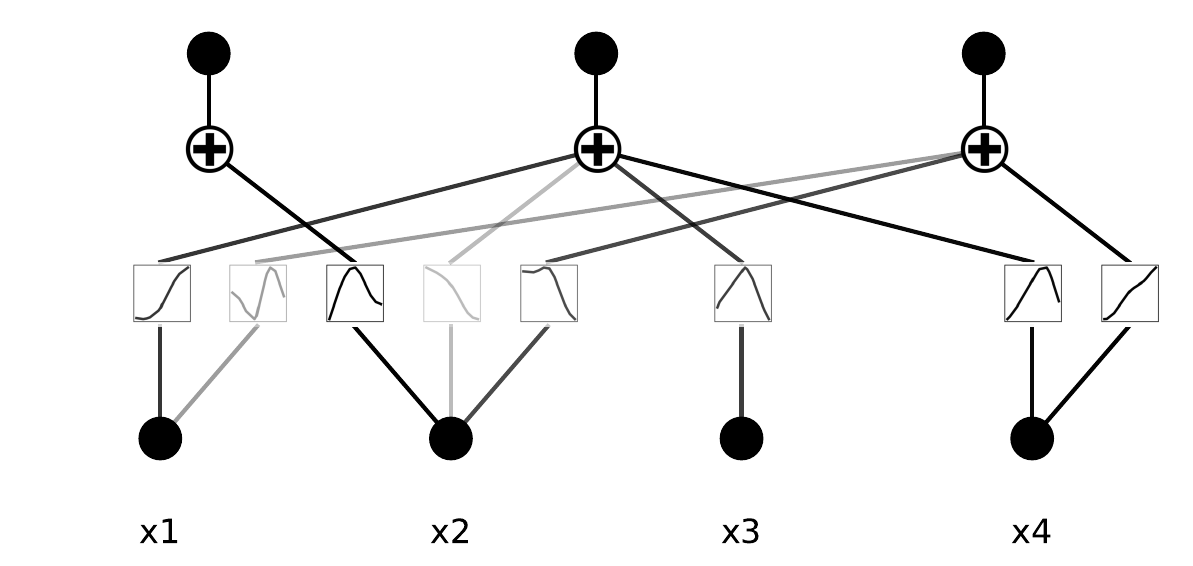}%
        \label{fig:structure_iris0}
    } \\
    \subfloat[Structure after fixed formula.]{%
        \includegraphics[width=0.8\linewidth]{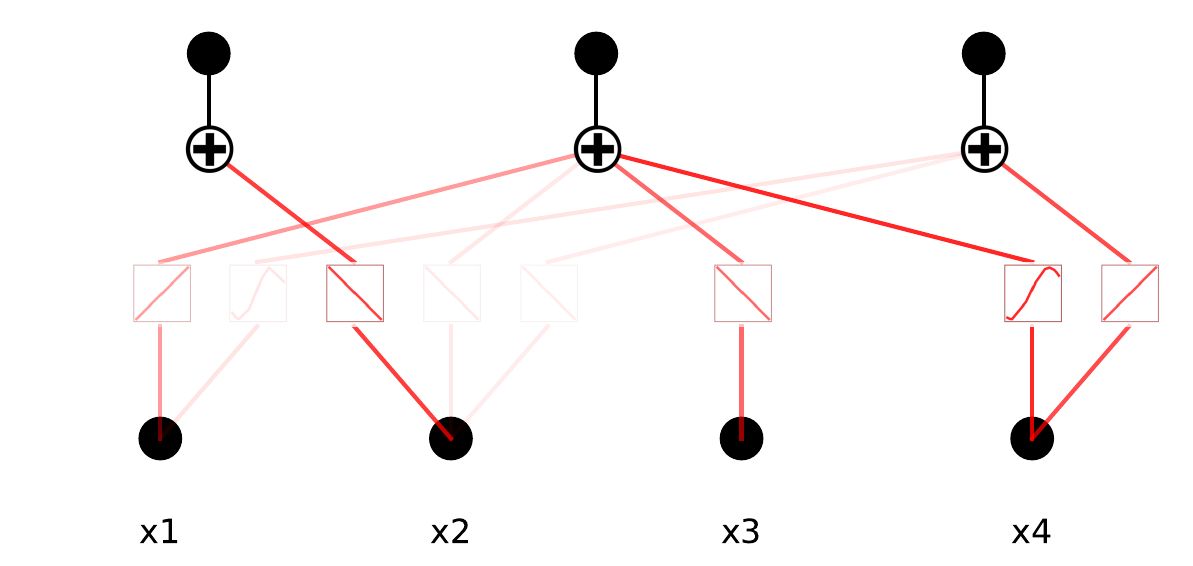}%
        \label{fig:structure_iris1}
    }
    \caption{Structural adjustments for enhanced interpretability on Iris dataset.}
    \label{fig:structure_iris}
\end{figure}

\subsection{Interpretability} \label{sec_Interpretability}

This section will demonstrate the interpretability of the experimental results using the Iris and Wine datasets as examples. GA-KAN optimizes the structure of KAN, aligning them more closely with the underlying distributions of the datasets. This optimization enhances the understanding of interpretability, allowing for clearer insights into how the models make decisions.

The interpretability of GA-KAN is achieved through three key aspects: feature selection, feature importance score, and symbolic formulae.
1) Feature selection, where GA-KAN encodes both network nodes and the connections between them, including features as input nodes. Throughout the evolutionary process, the encoding is optimized, allowing feature combinations that align more closely with the data distribution to achieve higher fitness scores during evaluation. This process effectively guides the model in selecting the most relevant features for the task.
2) Feature importance score, which provides a quantitative measure of each feature's contribution to the model's predictions. This can be visualized to reveal the influence of different features on the model's decision-making process and their relative importance in achieving accurate results.
3) Symbolic formulae, designed to be human-interpretable, allow a clear understanding of how the model calculates scores or probabilities for different categories.

\textbf{1) Iris Dataset:} Due to multiple network structures achieving the same performance in the final generation stage, the individual with fewer layers was selected as the representative example, the final result as shown in Fig.~\ref{fig:structure_iris}. Fig.~\ref{interp:iris} demonstrates the feature importance, represented by the attribution scores of the features. Fig.~\ref{fig:structure_iris0} illustrates the original structure of the network after completing the training process. The functional relationship between the category scores and the features can be observed. Initially, an attempt was made to automate the extraction process, but the results were suboptimal. Consequently, manual extraction was performed, and the formulae were derived based on the final structural figure, as mentioned earlier. Fig.~\ref{fig:structure_iris1} represents the result after converting the connections into corresponding formulae. The formulae extracted from the results are as follows:

\begin{subequations}\label{eqn-2}
  \footnotesize
  \begin{align}
    z_1 &= 59.08 - 34.97 x_{2} \\
    z_2 &= 376.42 x_{1} - 147.15 x_{2} - 288.7 x_{3} - 1295.19 \sin(1.95 x_{4} - 5.44 ) \nonumber \\
        &\quad - 116.67 \\
    z_3 &= - 247.23 x_{2} + 1953.46 x_{4} - 303.35 \sin(1.61 x_{1} - 0.28 ) - 1229.87
  \end{align}
\end{subequations}
here, $z_1$, $z_2$, and $z_3$ represent the raw scores for categories $1$, $2$, and $3$, respectively. Applying the softmax operation to $z$ yields the probabilities for each category, although this is not shown here. $x_1$ to $x_4$ represent the raw, unnormalized feature values. Eq.~\eqref{eqn-2} have been rounded to two decimal places.

From Fig~\ref{interp:iris}, it is evident that the first two features---\textit{petal width} and \textit{sepal width}---are sufficient for distinguishing the iris species. In contrast, the last two features---\textit{sepal length} and \textit{petal length}---contribute much less to the classification. Thus, focusing on the width features is generally more effective for identifying the species of an iris flower. Eq.~\eqref{eqn-2} demonstrate the relationships between the features and the category scores. For $z_1$, the relationships with the features are linear, whereas $z_2$ and $z_3$ exhibit nonlinear relationships involving sine functions. The nonlinear relationship indicates that the influence of the features on $z_2$ and $z_3$ is more complex. The sine functions introduce periodicity and nonlinearity, meaning that the effect of changes in the features on $z_2$ and $z_3$ varies depending on the values of the features, rather than remaining constant.

\begin{figure}[!t]
    \centering
        \includegraphics[width=0.98\linewidth]{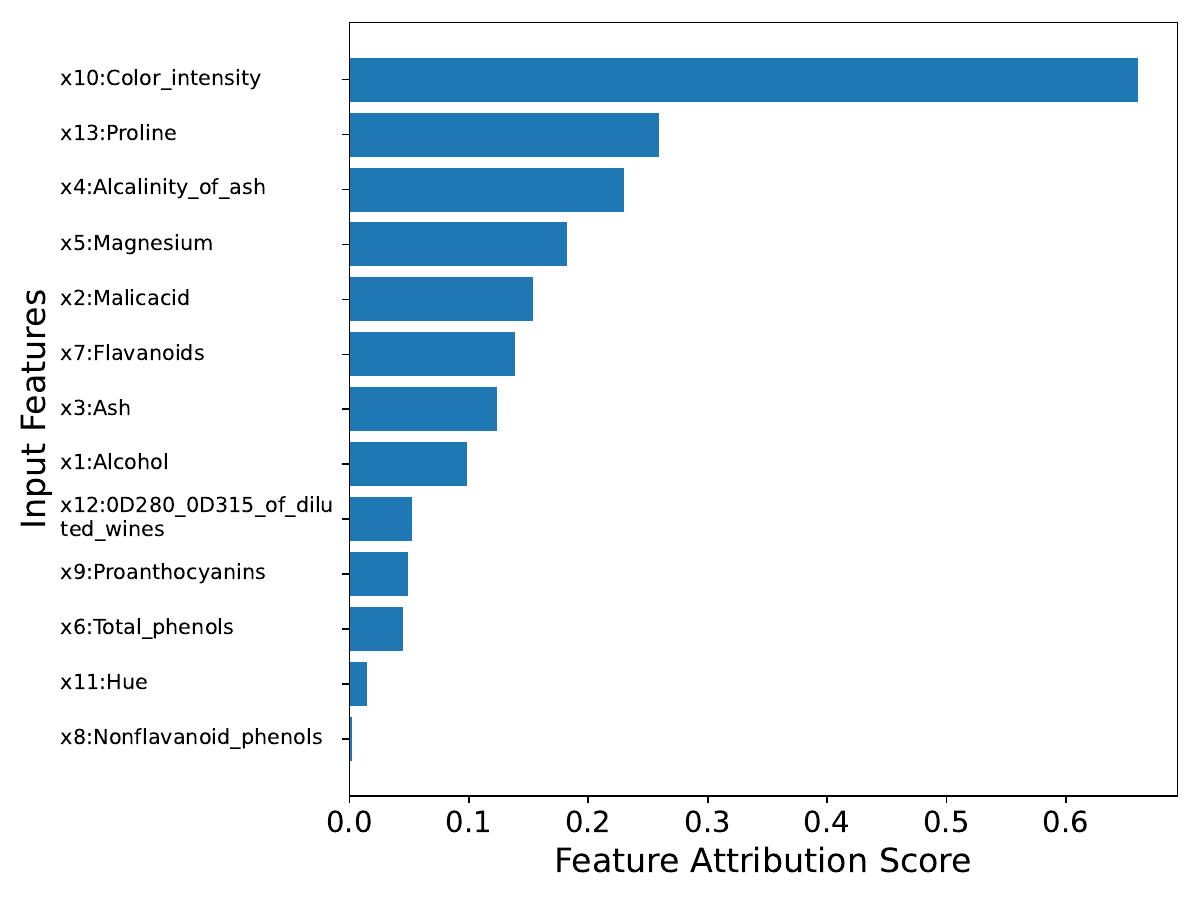}
    \caption{Feature attribution scores on Wine dataset.}
    \label{interp:wine}
\end{figure}

\begin{figure}[!t]
    \centering
    \subfloat[The original structure.]{%
        \includegraphics[width=0.95\linewidth]{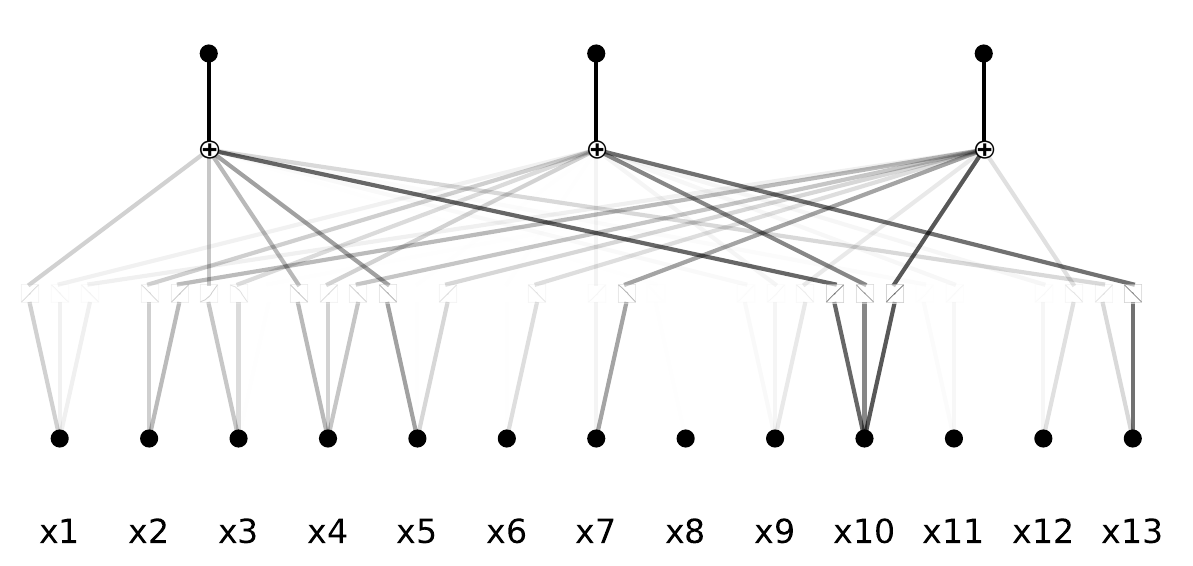}%
        \label{fig:structure_wine0}
    } \\
    \subfloat[Structure after fixed formula.]{%
        \includegraphics[width=0.95\linewidth]{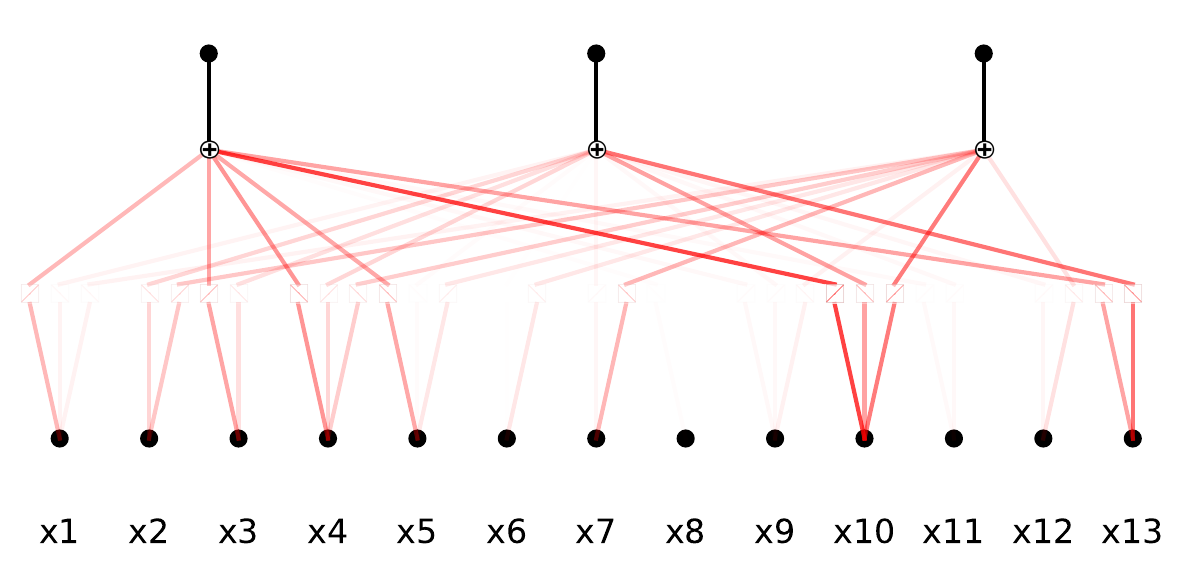}%
        \label{fig:structure_wine1}
    }
    \caption{Structural adjustments for enhanced interpretability on Wine dataset.}
    \label{fig:structure_wine}
\end{figure}

\textbf{2) Wine Dataset:} Similar to the Iris dataset, the interpretability results for the Wine dataset include both feature importance scores and symbolic formulae. The feature importance scores are presented in Fig.~\ref{interp:wine}, while the derived symbolic formulae are provided in Eq.~\eqref{eqn-3}. Additionally, the results of feature selection are also displayed.
As shown in Fig.~\ref{interp:wine}, the feature \textit{Color\_intensity} demonstrates the highest importance. This dominance is further highlighted in Fig.~\ref{fig:structure_wine0}, where the dark connection color for \textit{Color\_intensity} indicates its significant attribution. The results demonstrate that in the Wine dataset, the \textit{Color\_intensity} feature holds a dominant position.

Analysis of neural connections revealed that the relationships between functions are linear, leading to the simplified structure shown in Fig.~\ref{fig:structure_wine1}. The final formula derived from the network is presented in Eq.~\eqref{eqn-3}, as follows:

\begin{subequations}\label{eqn-3}
  \small
  \begin{align}
    z_1 &= 40.4 x_{1} + 45.1 x_{10} + 11.2 x_{11} - 0.1 x_{13} + 149.7 x_{3} \nonumber \\
        &\quad - 15.4 x_{4} - 3.1 x_{5} - 9.3 x_{8} + 6.2 x_{9} - 829.8 \\
    z_2 &= - 48.8 x_{1} - 134.0 x_{10} + 97.3 x_{11} + 37.2 x_{12} - 1.5 x_{13} \nonumber \\
        &\quad - 119.2 x_{2} - 379.6 x_{3} + 41.4 x_{4} - 1.1 x_{5} + 2.9 x_{6} + 31.8 x_{7} \nonumber \\
        &\quad + 3.1 x_{8} + 43.7 x_{9} + 1971.9 \\
    z_3 &= - 25.2 x_{1} + 97.0 x_{10} - 62.8 x_{12} + 78.9 x_{2} + 2.7 x_{3} - 24.9 x_{4} \nonumber \\
        &\quad + 3.0 x_{5} - 67.2 x_{6} - 123.4 x_{7} - 46.4 x_{9} - 48.4
  \end{align}
\end{subequations}
where, $z_1$ to $z_3$ represent the scores for categories $1$ to $3$, respectively. Applying the softmax operation to $z$ yields the probabilities for each category, although this is not shown here. $x_1$ to $x_{13}$ represent the raw, unnormalized feature values. Eq.~\eqref{eqn-3} have been rounded to two decimal places. However, this equation does not include feature normalization, meaning each feature is computed in its original scale. During computation, the values of the features can be directly substituted without the need for data transformation, making the process clear and straightforward. Nevertheless, due to the lack of normalization, Eq.~\eqref{eqn-3} may not be suitable for comparing features with different scales. Therefore, normalization was applied, resulting in the adjusted formulae shown in Eq.~\eqref{eqn-4}. From Eq.~\eqref{eqn-4}, it is evident that features with larger coefficients have a greater impact on the category scores, while the signs of the coefficients indicate positive or negative influences. Eq.~\eqref{eqn-3} and Eq.~\eqref{eqn-4} are simpler than those derived from the Iris dataset. This simplicity stems from the linear relationship between the scores and the features, with only first-degree terms of $x$ involved.

\begin{subequations}\label{eqn-4}
  \small
  \begin{align}
    z_1 &= 33.3 x_{1} + 97.3 x_{10} + 2.6 x_{11} - 38.9 x_{13} + 40.7 x_{3} \nonumber \\
        &\quad - 52.6 x_{4} - 45.6 x_{5} - 1.2 x_{8} + 3.6 x_{9} - 415.3 \\
    z_2 &= - 40.2 x_{1} - 288.8 x_{10} + 22.5 x_{11} + 26.2 x_{12} - 467.0 x_{13} \nonumber \\
        &\quad - 128.6 x_{2} - 103.2 x_{3} + 141.5 x_{4} - 16.3 x_{5} + 1.8 x_{6} + 32.2 x_{7} \nonumber \\
        &\quad + 0.4 x_{8} + 25.5 x_{9} - 587.9 \\ 
    z_3 &= - 20.8 x_{1} + 209.1 x_{10} - 44.2 x_{12} + 85.2 x_{2} + 0.7 x_{3} - 85.2 x_{4} \nonumber \\
        &\quad + 43.9 x_{5} - 42.6 x_{6} - 125.1 x_{7} - 27.1 x_{9} - 538.4
  \end{align}
\end{subequations}

These formulae enhance the transparency of the model's decision-making process, enabling humans to understand and interpret how predictions are made based on input variables. From a feature selection perspective, GA-KAN effectively excludes the feature $x_8$ (representing \textit{Nonflavanoid\_phenols}) on the Wine dataset, as confirmed by its absence in Eq.~\eqref{eqn-3}. This feature exclusion demonstrates GA-KAN's ability to perform feature selection, facilitating the design of a more compact network to address the task, thereby reducing parameter consumption and improving efficiency.
By employing a wrapper-based approach for feature selection \cite{jiao2023survey}, GA-KAN iteratively evaluates different feature subsets through model training, identifying the most relevant features to optimize both model structure and performance. This process reduces redundancy and enhances the model's interpretability and efficiency by focusing on feature combinations that contribute most to decision-making.

\section{Conclusion and Future Work}\label{sec:conclusion_future_work}

In this paper, the overall goal of the proposed GA-KAN---a genetic algorithm-based framework for the automatic optimization of KANs---has been achieved through four specific contributions. Firstly, a new encoding strategy was implemented, encoding the neuron connections, grid values, and depth of KANs into chromosomes. Secondly, a new decoding process was developed, incorporating a \textit{degradation mechanism} and a zero mask technique, which enables more efficient exploration of diverse KAN configurations. Thirdly, GA-KAN automatically optimizes both the structure and grid values of KANs, requiring no human intervention in the design process and minimal adjustment in the formula extraction process. Lastly, the accuracy, interpretability, and parameter reduction of GA-KAN were validated across multiple experiments. Specifically, GA-KAN was validated using two toy datasets from the original KAN paper, eliminating the need for manual parameter adjustments. GA-KAN achieved 100\% accuracy on the Wine, Iris, and WDBC datasets, 90.00\% on Raisin, and 95.14\% on Rice, surpassing traditional models and outperforming the standard KAN proposed in the original paper. Furthermore, GA-KAN significantly reduced the number of parameters across all five datasets. Additionally, the results for the Wine and Iris datasets provided symbolic formulae, further demonstrating the enhanced interpretability of the model.

Due to the efficiency limitations of KAN and the high computational cost of NAS, we validated it only on relatively smaller-scale datasets.
Looking ahead, future work can expand GA-KAN to other types of tasks like regression to demonstrate its versatility and robustness. Furthermore, exploring strategies for deploying GA-KAN on resource-constrained hardware and scaling it to optimize larger and more complex systems, including both neural networks and datasets, will be essential for enhancing its practical utility. In addition, future work might also need to focus on applying GA-KAN to more challenging problems to further assess its performance. While GA-KAN offers a compelling framework for neural architecture search and shows significant potential, addressing computational efficiency and hardware optimization will be key for its practical use in demanding environments.

\bibliographystyle{IEEEtran}
\bibliography{references}

\vfill

\end{document}